\documentclass[twoside,11pt]{article}

\usepackage{amsmath,bm,paralist}
\usepackage{jmlr2eArxiv}
\usepackage{algorithm}
\usepackage{algorithmic}

\def \R {\mathbb{R}}
\def \D {\mathcal{D}}
\def \y {\mathbf{y}}

\def \x {\mathbf{x}}

\def \L {\mathcal{L}}
\def \G {\mathcal{G}}

\def \v {\mathbf{v}}

\def \S {\mathcal{S}}

\def \z {\mathbf{z}}

\def \Lh {\widehat{\L}}
\def \Gh {\widehat{\G}}
\def \Gt {\widetilde{\G}}
\def \Lt {\widetilde{\L}}

\def \u {\mathbf{u}}

\def \v {\mathbf{v}}
\def \w {\mathbf{w}}

\def \R {\mathbb{R}}

\def \G {\mathcal{G}}

\def \Ah {\widehat{A}}

\def \e {\mathbf{e}}

\def \Ah {\widehat{A}}

\def \N {\mathcal{N}}
\def \wh {\widehat{\w}}

\def \K {\mathcal{K}}
\def \lt {\widetilde{\lambda}}

\def \wh {\widehat{\w}}
\def \Ol {\Omega_{\lambda}}
\def \Ow {\Omega_{w}}
\def \Obl {\bar{\Omega}_{\lambda}}
\def \Obw {\bar{\Omega}_w}
\def \wt {\widetilde{\w}}

\def \lbd {\bm \lambda}
\def \sgn {\mbox{sign}}

\def \lt {\widetilde{\lbd}}
\def \lh {\widehat{\lbd}}

\DeclareMathOperator*{\conv}{conv}

\newtheorem{thm}{Theorem}

\newtheorem{lem}{Lemma}
\newtheorem{cor}{Corollary}
\newtheorem{property}{Property}

\jmlrheading{1}{2015}{1-48}{9/15}{}{Lijun Zhang, Tianbao Yang, Rong Jin, and Zhi-Hua Zhou}

\ShortHeadings{Sparse Learning for Large-scale and High-dimensional Data}{Zhang, Yang, Jin, and Zhou}
\firstpageno{1}

\begin{document}

\title{Sparse Learning for Large-scale and High-dimensional Data: A Randomized Convex-concave Optimization Approach}

\author{\name Lijun Zhang \email zhanglj@lamda.nju.edu.cn\\
       \addr National Key Laboratory for Novel Software Technology\\
       Nanjing University, Nanjing 210023, China
       \AND
       \name Tianbao Yang \email tianbao-yang@uiowa.edu\\
       \addr Department of Computer Science\\
       the University of Iowa, Iowa City, IA 52242, USA
       \AND
       \name Rong Jin \email rongjin@cse.msu.edu\\
       \addr Department of Computer Science and Engineering\\
        Michigan State University, East Lansing, MI 48824, USA
       \AND
       \name Zhi-Hua Zhou \email zhouzh@lamda.nju.edu.cn\\
       \addr National Key Laboratory for Novel Software Technology\\
       Nanjing University, Nanjing 210023, China}

\editor{Leslie Pack Kaelbling}

\maketitle

\begin{abstract}
In this paper,  we develop a randomized algorithm and theory for learning a sparse model from large-scale and high-dimensional data, which is usually formulated as an empirical risk minimization problem with a sparsity-inducing regularizer. Under the assumption that there exists a (approximately) sparse solution with high classification accuracy, we argue that the dual solution is also sparse or approximately sparse. The fact that both primal and dual solutions are sparse motivates us to develop a randomized approach for a general convex-concave optimization problem. Specifically, the proposed approach combines the strength of random projection with that of sparse learning: it utilizes random projection to reduce the dimensionality, and introduces $\ell_1$-norm regularization to alleviate the approximation error caused by random projection. Theoretical analysis shows that under favored conditions, the randomized algorithm can accurately recover the optimal solutions to the convex-concave optimization problem (i.e., recover both the primal and dual solutions).
\end{abstract}


\section{Introduction}
Learning the sparse representation of a predictive model has received considerable attention in recent years~\citep{Bach:2012:OSP}. Given a set of training examples $\{(\x_i,\y_i)\}_{i=1}^n$ with $\x_i \in \R^d$ and $y_i \in \R$, the optimization problem is generally formulated as
\begin{equation} \label{eqn:supervised}
\min_{\w \in \Omega} \frac{1}{n} \sum_{i=1}^n \ell( y_i\x_i^\top \w) +\gamma \psi(\w)
\end{equation}
where $\ell(\cdot)$ is a convex function such as the logistic loss to measure the empirical error, and $\psi(\cdot)$ is a sparsity-inducing regularizer such as the elastic net \citep{RSSB:RSSB} to avoid overfitting~\citep{The-Elements-of-Statistical-Learning-2009}. When both $d$ and $n$ are very large, directly solving (\ref{eqn:supervised}) could be computationally expensive. A straightforward way to address this challenge is first reducing the dimensionality of the data, then solving a low-dimensional problem, and finally mapping the solution back to the original space. The limitation of this approach is that the final solution, after mapping from the low-dimensional space to the original high-dimensional space, may not be sparse.

The goal of this paper is to develop an efficient algorithm for solving the problem in (\ref{eqn:supervised}), and at the same time preserve the (approximate) sparsity of the solution. Our approach is motivated by the following simple observation:
\begin{quote}
     If there exists a sparse model with high prediction accuracy, the dual solution to (\ref{eqn:supervised}) is also sparse or approximately sparse.
\end{quote}
To see this, let us formulate (\ref{eqn:supervised}) as a convex-concave optimization problem. By writing $\ell(z)$ in its convex conjugate form, i.e.,
\[
    \ell(z) = \max\limits_{\lambda \in \Gamma} \lambda z - \ell_*(\lambda),
\]
where $\ell_*(\cdot)$ is the Fenchel conjugate of $\ell(\cdot)$~\citep{Convex:analysis} and $\Gamma$ is the domain of the dual variable, we get the following convex-concave formulation:
\begin{equation} \label{eqn:conv:conca}
\max\limits_{\lbd \in \Gamma^n} \min\limits_{\w \in \Omega} \; \gamma n \psi(\w)   - \sum_{i=1}^n \ell_*(\lambda_i) +  \sum_{i=1}^n \lambda_i y_i \x_i^\top \w.
\end{equation}
Denote the optimal solutions to (\ref{eqn:conv:conca}) by $(\w_*, \lbd_*)$. By the Fenchel conjugate theory~\cite[Lemma 11.4]{bianchi-2006-prediction}, we have
\[
[\lbd_*]_i= \ell'( y_i\x_i^\top \w_*).
\]
Let us consider the squared hinge loss for classification~\citep{Tsochantaridis:2005:LMM}, where $\ell(z)=\max(0, 1-z)^2$. Therefore, $y_i\x_i^\top \w_* \geq 1$ indicates that $[\lbd_*]_i=0$. As a result, when most of the examples can be classified by a large margin (which is likely to occur in large-scale and high-dimensional setting), it is reasonable to assume that the dual solution is sparse. Similarly, for logistic regression, we can argue the dual solution is approximately sparse.

Abstracting (\ref{eqn:conv:conca}) slightly, in the following, we will study a general convex-concave optimization problem:
\begin{equation} \label{eqn:cc-opt}
\max\limits_{\lbd \in \Delta} \min\limits_{\w \in \Omega} \; g(\w) - h(\lbd) - \w^{\top}A\lbd
\end{equation}
where $\Delta \subseteq \R^n$ and $\Omega \subseteq \R^d$ are the domains for $\lbd$ and $\w$, respectively, $g(\cdot)$ and $h(\cdot)$ are two convex functions, and $A \in \R^{d\times n}$ is a matrix. The benefit of analyzing (\ref{eqn:cc-opt}) instead of (\ref{eqn:supervised}) is that the convex-concave formulation allows us to exploit the prior knowledge that \emph{both} $\w_*$ and $\lbd_*$ are sparse or approximately sparse. The problem in (\ref{eqn:cc-opt}) has been widely studied in the optimization community, and when $n$ and $d$ are medium size, it can be solved iteratively by gradient based methods~\citep{Nesterov_Non_Smooth,nemirovski-2005-prox}.

We assume the two convex functions $g(\cdot)$ and $h(\cdot)$ are relatively simple such that evaluating their values or gradients takes $O(d)$ and $O(n)$ complexities, respectively. The bottleneck is the computations involving the bilinear term $\w^{\top}A\lbd$, which have $O(nd)$ complexity in both time and space. To overcome this difficulty, we develop a randomized algorithm that solves (\ref{eqn:cc-opt}) approximately but at a significantly lower cost. The proposed algorithm combines two well-known techniques---\emph{random projection} and \emph{$\ell_1$-norm regularization} in a principled way. Specifically, random projection is used to find a low-rank approximation of $A$, which not only reduces the storage requirement but also accelerates the computations. The role of $\ell_1$-norm regularization is twofold. One one hand, it is introduced to compensate for the distortion caused by randomization, and on the other hand it enforces the sparsity of the final solutions. Under mild assumptions about the optimization problem in (\ref{eqn:cc-opt}), the proposed algorithm has a small recovery error provided the optimal solutions to (\ref{eqn:cc-opt}) are sparse or approximately sparse.

\section{Related Work}
Random projection has been widely used as an efficient algorithm for dimensionality reduction~\citep{kaski-1998-dimensionality,bingham-2001-random}. In the case of unsupervised learning, it has been proved that random projection is able to preserve the distance~\citep{RSA:RSA}, inner product~\citep{ML06:Arriaga}, volumes and distance to affine spaces~\citep{RP:Volumes}. In the case of supervised learning, random projection is generally used as a preprocessing step to find a low-dimensional representation of the data, and thus reduces the computational cost of training. For classification, theoretical studies mainly focus on examining the generalization error or the preservation of classification margin in the low-dimensional space~\citep{ML06:Balcan,Margin_RP,RP:SVM}. For regression, there do exist theoretical guarantees for the recovery error, but they only hold for the least squares problem~\citep{Mahoney:2011:RAM}.

Our work is closely related to Dual Random Projection (DRP)~\citep{COLT13:Zhang,TIT:2014:Zhang} and Dual-sparse Regularized Randomized Reduction (DSRR)~\citep{Dual:Sparse}, which also investigate random projection from the perspective of optimization. However, both DRP and DSRR are limited to the special case that $\psi(\w)=\|\w\|_2^2$, which leads to a simple dual problem. In contrast, our algorithm is designed for the case that $\psi(\cdot)$ is a sparsity-inducing regularizer, and built upon the convex-concave formulation. Similar to DSRR, our algorithm makes use of the sparsity of the dual solution, but we further exploit the sparsity of the primal solution. A noticeable advantage of our analysis is the mild assumption about the data matrix $A$. To recover the primal solution, DRP assumes the data matrix is low-rank and DSRR assumes it satisfies the restricted eigenvalue condition, in contrast, our algorithm only requires columns or rows of $A$ are bounded.

There are many literatures that study the statistical property of the sparse learning problem in (\ref{eqn:supervised}) \citep{Random:Projection,agarwal2012,ICML2012Xiao,AISTATS:2015:Zhang}. For example, in the context of compressive sensing \citep{Intro_CS}, it has been established that a sparse signal can be recovered up to an $O(\sqrt{s \log d /n})$ error, where $s$ is the sparsity of the unknown signal. We note that the statistical error is not directly comparable to the optimization error derived in this paper. That is because the analysis of statistical error relies on heavy assumptions about the data, e.g., the RIP condition \citep{Candes:RIP}. On the other hand, the optimization error is derived under very weak conditions.

\section{Algorithm}
To reduce the computational cost of (\ref{eqn:cc-opt}), we first generate a random matrix $R\in \R^{n\times m}$, where $m \ll \min(d,n)$.  Define $\Ah =  A R \in \R^{d \times m}$, we propose to solve the following problem
\begin{equation}\label{eqn:cc-opt-1}
\max\limits_{\lbd \in \Delta} \min\limits_{\w \in \Omega} \; g(\w) - h(\lbd) - \w^{\top}\Ah R^{\top}\lbd + \gamma_{w}\|\w\|_1 - \gamma_{\lambda} \|\lbd\|_1
\end{equation}
where $\gamma_w$ and $\gamma_\lambda$ are two regularization parameters. The construction of the random matrix $R$, as well as the values of the two regularization parameters $\gamma_{w}$ and $\gamma_{\lambda}$ will be discussed later.  The optimization problem in (\ref{eqn:cc-opt-1}) can be solved by algorithms designed for composite convex-concave problems \citep{Primal_Dual,Composite:Convex:Concave}.

Compared to (\ref{eqn:cc-opt}), the main advantage of (\ref{eqn:cc-opt-1}) is that it only needs to load $\Ah$ and $R$ into the memory, making it convenient to deal with large-scale problems. With the help of random projection, the computational complexity for evaluating the value and gradient is reduced from $O(dn)$ to $O(dm+nm)$. Compared to previous randomized algorithms~\citep{ML06:Balcan,COLT13:Zhang,Dual:Sparse}, (\ref{eqn:cc-opt-1}) has two new features: i) the optimization is still performed in the original space; and ii) the $\ell_1$-norm is introduced to regularize both primal and dual solutions. As we will prove later, the combination of these two features will ensure the solutions to (\ref{eqn:cc-opt-1}) are approximately sparse. Finally, note that in (\ref{eqn:cc-opt-1}) $RR^\top$ is inserted at the right side of $A$, it can also be put at the left side of $A$. In this case, we have the following optimization problem
\begin{equation} \label{eqn:for:lambda}
\max\limits_{\lbd \in \Delta} \min\limits_{\w \in \Omega} \; g(\w) - h(\lbd) - \w^{\top} R \Ah \lbd + \gamma_{w}\|\w\|_1 - \gamma_{\lambda} \|\lbd\|_1
\end{equation}
where $R\in \R^{d \times m}$ is a random matrix, and $\Ah =  R^\top A \in \R^{m \times n}$.

Let $(\w_*, \lbd_*)$ and $(\wh, \lh)$ be the optimal solution to the convex-concave optimization problem in (\ref{eqn:cc-opt}) and (\ref{eqn:cc-opt-1})/(\ref{eqn:for:lambda}), respectively. Under suitable conditions, we will show that
\[
\begin{split}
 \|\wh - \w_*\|_2  & \leq O\left(\sqrt{ \frac{\|\w_*\|_0 \|\lbd_*\|_0 \log n}{m}} \right)  \textrm{ and } \\
  \|\lh - \lbd_*\|_2 & \leq O\left(\sqrt{ \frac{\|\w_*\|_0 \|\lbd_*\|_0 \log d}{m}} \right)
 \end{split}
\]
implying a small recovery error when $\w_*$ and $\lbd_*$ are sparse. A similar recovery guarantee also holds when the optimal solutions to (\ref{eqn:cc-opt}) are approximately sparse, i.e., when they can be well-approximated by sparse vectors.

\section{Main Results}
We first introduce common assumptions that we make, and then present theoretical guarantees.
\subsection{Assumptions}
\paragraph{Assumptions about (\ref{eqn:cc-opt})}
We make the following assumptions about  (\ref{eqn:cc-opt}).
\begin{compactitem}
   \item $g(\w)$ is $\alpha$-strongly convex with respect to the Euclidean norm.  Let's take the optimization problem in (\ref{eqn:conv:conca}) as an example. (\ref{eqn:conv:conca}) will satisfy this assumption if some strongly convex function (e.g., $\|\w\|_2^2$) is a part of the regularizer $\psi(\w)$.
   \item $h(\lbd)$ is $\beta$-strongly convex with respect to the Euclidean norm. For the problem in (\ref{eqn:conv:conca}), if $\ell(\cdot)$ is a smooth function (e.g., the logistic loss), then its convex conjugate $\ell_*(\cdot)$ will be strongly convex~\citep{Convex:analysis,Duality:convex:smooth}.
   \item Either columns or rows of $A$ have bounded $\ell_2$-norm. Without loss of generality, we assume
   \begin{align}
   \|A_{i*}\|_2 &\leq 1, \ \forall  i \in[d], \label{eqn:upper:A1}\\
    \|A_{*j}\|_2& \leq 1,\ \forall j \in [n] \label{eqn:upper:A2}.
   \end{align}
    The above assumption can be satisfied by normalizing rows or columns of $A$. 
\end{compactitem}

\paragraph{Assumptions about $R$} We assume the random matrix $R \in \R^{n\times m}$ has the following property.
\begin{compactitem}
  \item With a high probability, the linear operator $R^\top: \R^n\mapsto\R^m$ is able to preserve the $\ell_2$-norm of its input. In mathematical terms, we need the following property.
  \begin{property} \label{thm:jl} There exists a constant $c>0$, such that
\[
    \Pr\left\{(1 - \varepsilon)\|\x\|_2^2 \leq \|R^\top \x\|_2^2 \leq (1 + \varepsilon)\|\x\|_2^2 \right\} \geq 1 - 2\exp\left(- \frac{m \varepsilon^2}{c} \right)
\]
for any fixed $\x \in \R^d$ and $0<\epsilon \leq 1/2$.
\end{property}
\end{compactitem}
The above property is widely used to prove the famous Johnson--Lindenstrauss lemma~\citep{RSA:RSA}. Let $R=\frac{1}{\sqrt{m}} S$. Previous studies~\citep{JL_Database,ML06:Arriaga} have proved that Property~\ref{thm:jl} is true if $\{S_{ij}\}$ are independent random variables sampled from the Gaussian distribution $\N(0,1)$, uniform distribution over $\{\pm 1\}$, or the following database-friendly distribution
\[
X= \left\{
     \begin{array}{ll}
       \sqrt{3}, & \hbox{with probability } 1/6; \\
       0, & \hbox{with probability } 2/3; \\
       -\sqrt{3}, & \hbox{with probability } 1/6.
     \end{array}
   \right.
\]
More generally, a sufficient condition for Property~\ref{thm:jl} is that columns of $R$ are independent, isotropic, and subgaussian vectors~\citep{UUP}.
\subsection{Theoretical Guarantees}
\subsubsection{Sparse Solutions}
We first consider the case that both $\w_*$ and $\lbd_*$ are sparse. Define
\[
s_w = \|\w_*\|_0, \textrm{ and } s_{\lambda} = \|\lbd_*\|_0.
\]
We have the following theorem for the optimization problem in (\ref{eqn:cc-opt-1}).

\begin{thm} \label{thm:main}
Let  $(\wh, \lh)$ be the optimal solution to the problem in  (\ref{eqn:cc-opt-1}). Set
\begin{align}
\gamma_{\lambda} &\geq 2 \|A^\top\w_*\|_2 \sqrt{\frac{c}{m}\log \frac{4 n}{\delta}}, \label{eqn:gamma:lambda}\\
\gamma_{w} &\geq 2 \|\lbd_*\|_2 \sqrt{\frac{c}{m}\log \frac{4d}{\delta}} + \frac{6\gamma_\lambda\sqrt{s_\lambda}}{\beta} \left(1+7 \sqrt{\frac{c}{m}\left(\log \frac{4d}{\delta} +  16s_\lambda \log \frac{9 n}{8s_\lambda}\right)}\right)\label{eqn:gamma:w}.
\end{align}
With a probability at least $1-3\delta$, we have
\[
\|\wh - \w_*\|_2 \leq \frac{3\gamma_w\sqrt{s_w}}{\alpha}, \ \|\wh - \w_*\|_1 \leq \frac{12\gamma_w s_w}{\alpha}, \textrm{ and } \frac{\|\wh - \w_*\|_1}{\|\wh - \w_*\|_2} \leq 4\sqrt{s_w}
\]
provided
\begin{equation} \label{eqn:delta}
m \geq 4c \log \frac{4}{\delta}
\end{equation}
where $c$ is the constant in Property~\ref{thm:jl}.
\end{thm}
Notice that $\|\wh - \w_*\|_1/\|\wh - \w_*\|_2 \leq 4\sqrt{s_w}$ indicates that $\wh-\w_*$ is approximately sparse~\citep{OneBit:Plan:LP,OneBit:Plan:Robust}. Combining with the fact $\w_*$ is sparse, we conclude that $\wh$ is also approximately sparse.

Then, we discuss the recovery guarantee for the sparse learning problem in (\ref{eqn:supervised}) or (\ref{eqn:conv:conca}). Since $A^\top\w_* \in \R^n$, we can take $\|A^\top\w_*\|_2 = O(\sqrt{n})$. Since $\|\lbd_*\|_0=s_{\lambda}$, we can assume  $\|\lbd_*\|_2=O(\sqrt{s_{\lambda}})$.  According to the theoretical analysis of regularized empirical risk minimization~\citep{Wu:2005:SSM,NIPS2008_3400,Oracle_inequality}, the optimal $\gamma$, that minimizes the generalization error, can be chosen as $\gamma=O(1/\sqrt{n})$, and thus $\alpha=O(\gamma n)=O(\sqrt{n})$. When the loss $\ell(\cdot)$ is smooth, we have $\beta=O(1)$.  The following corollary provides a simplified result based on the above discussions.
\begin{cor} \label{cor:1}
Assume $\|A^\top\w_*\|_2 = O(\sqrt{n})$, $\|\lbd_*\|_2=O(\sqrt{s_{\lambda}})$, $\alpha=O(\sqrt{n})$, and $\beta=O(1)$. When $m \geq O(s_\lambda \log  n)$, we can choose
\[
\gamma_{\lambda} = O\left(\sqrt{\frac{n \log n}{m}}  \right) \textrm{ and }  \gamma_{w} =  O\left(\sqrt{\frac{s_{\lambda} \log d}{m}}+  \gamma_{\lambda} \sqrt{s_\lambda}  \right) = O\left(\sqrt{ \frac{n s_\lambda \log n}{m}}  \right)
\]
such that with a high probability
\[
\|\wh - \w_*\|_2 \leq O\left( \frac{\gamma_w\sqrt{s_w}}{\sqrt{n}} \right) = O\left(\sqrt{ \frac{s_w s_\lambda \log n}{m}}  \right) \textrm{ and } \frac{\|\wh - \w_*\|_1}{\|\wh - \w_*\|_2} \leq 4\sqrt{s_w}.
\]
\end{cor}

A natural question to ask is whether similar recovery guarantees for $\lh$ can be proved under the conditions in Theorem~\ref{thm:main}. Unfortunately, we are not able to give a positive answer, and only have the following theorem.
\begin{thm} \label{thm:2}
Assume $\gamma_{\lambda}$ satisfies the condition in (\ref{eqn:gamma:lambda}). With a probability at least $1-\delta$, we have
\[
\|\lh - \lbd_*\|_2 \leq  \frac{3\gamma_{\lambda} \sqrt{s_\lambda}}{\beta}+  \frac{2}{\beta} \left( 1+ \|RR^{\top}-I\|_2\right)\|A^{\top}(\wh-\w_*)\|_2
\]
provided (\ref{eqn:delta}) holds.
\end{thm}
The upper bound in the above theorem is quite loose, because $\|RR^{\top}-I\|_2$ is roughly on the order of $n \log n/m$~\citep{Sum_Matrix}.

Due to the symmetry between $\lbd$ and $\w$, we can recover $\lbd_*$  via  (\ref{eqn:for:lambda}) instead of (\ref{eqn:cc-opt-1}). Then, by
replacing $\w_*$ in Theorem~\ref{thm:main} with $\lbd_*$, $\wh$ with $\lh$, $n$ with $d$, and so on, we obtain the following theoretical guarantee.
\begin{thm} \label{thm:main:lambda}
Let  $(\wh, \lh)$ be the optimal solution to the problem in (\ref{eqn:for:lambda}). Set
\begin{align*}
\gamma_{w} &\geq 2 \|A \lbd_*\|_2 \sqrt{\frac{c}{m}\log \frac{4 d}{\delta}}, \\
\gamma_{\lambda} &\geq 2 \|\w_*\|_2 \sqrt{\frac{c}{m}\log \frac{4n}{\delta}} + \frac{6\gamma_w \sqrt{s_w}}{\alpha} \left(1+7 \sqrt{\frac{c}{m}\left(\log \frac{4n}{\delta} +  16s_w \log \frac{9 d}{8s_w}\right)}\right).
\end{align*}
With a probability at least $1-3\delta$, we have
\[
\|\lh - \lbd_*\|_2 \leq \frac{3\gamma_\lambda\sqrt{s_\lambda}}{\beta}, \ \|\lh - \lbd_*\|_1 \leq \frac{12\gamma_\lambda s_\lambda}{\beta}, \textrm{ and } \frac{\|\lh - \lbd_*\|_1}{\|\lh - \lbd_*\|_2} \leq 4\sqrt{s_\lambda}
\]
provided (\ref{eqn:delta}) holds.
\end{thm}

To simplify the above theorem, we can take $\|A \lbd_*\|_2 =O(\sqrt{d})$ since $A \lbd_* \in \R^d$. Because (\ref{eqn:supervised}) has both a constraint and a regularizer, we can assume the optimal primal solution is well-bounded, that is, $\|\w_*\|_2=O(1)$. Finally, we assume $d\leq O(n)$, and have the following corollary.
\begin{cor}
Assume $\|A \lbd_*\|_2= O(\sqrt{d})$, $\|\w_*\|_2=O(1)$, $\alpha=O(\sqrt{n})$, $\beta=O(1)$, and $d \leq O(n)$. When $m \geq O(s_w \log  d)$, we can choose
\[
\gamma_{w} = O\left(\sqrt{\frac{d \log d}{m}}  \right) \textrm{ and }  \gamma_{\lambda} =  O\left(\sqrt{\frac{\log n}{m}}+  \gamma_{w} \sqrt{\frac{s_w}{n}}  \right) \leq O\left(\sqrt{ \frac{s_w \log d}{m}}  \right)
\]
such that with a high probability
\[
\|\lh - \lbd_*\|_2 \leq O\left(\gamma_\lambda\sqrt{s_\lambda}\right)= O\left(\sqrt{ \frac{s_w s_\lambda \log d}{m}}  \right) \textrm{ and }  \frac{\|\lh - \lbd_*\|_1}{\|\lh - \lbd_*\|_2} \leq  4 \sqrt{ s_\lambda}.
\]
\end{cor}

\subsubsection{Approximately Sparse Solutions} \label{sec:app:sparse}
We now proceed to study the case that the optimal solutions to (\ref{eqn:cc-opt}) are only approximately sparse.

With a slight abuse of notation, we assume $\w_*$ and $\lbd_*$ are two sparse vectors, with $\|\w_*\|_0 = s_w$ and $\|\lbd_*\|_0 = s_{\lambda}$, that solve (\ref{eqn:cc-opt}) approximately in the sense that
\begin{align}
\| \nabla g(\w_*)-  A \lbd_* \|_\infty& \leq \varsigma, \label{eqn:app:second:1}\\
\|\nabla h(\lbd_*) + A^{\top}\w_*\|_{\infty} &\leq \varsigma, \label{eqn:app:second:2}
\end{align}
for some small constant $\varsigma>0$. The above conditions can be considered as sub-optimality conditions~\citep{Convex-Optimization} of $\w_*$ and $\lbd_*$ measured in the $\ell_\infty$-norm. After a similar analysis, we have the following theorem.
\begin{thm} \label{thm:new:approximate} Let  $(\wh, \lh)$ be the optimal solution to the problem in (\ref{eqn:cc-opt-1}). Assume (\ref{eqn:app:second:1}) and (\ref{eqn:app:second:2}) hold. Set
\begin{align*}
\gamma_{\lambda} &\geq 2 \|A^\top\w_*\|_2 \sqrt{\frac{c}{m}\log \frac{4 n}{\delta}} + 2\varsigma,\\
\gamma_{w} &\geq 2 \|\lbd_*\|_2 \sqrt{\frac{c}{m}\log \frac{4d}{\delta}} + \frac{6\gamma_\lambda\sqrt{s_\lambda}}{\beta} \left(1+7 \sqrt{\frac{c}{m}\left(\log \frac{4d}{\delta} +  16s_\lambda \log \frac{9 n}{8s_\lambda}\right)}\right)+ 2 \varsigma.
\end{align*}
With a probability at least $1-3\delta$, we have
\[
\|\wh - \w_*\|_2 \leq \frac{3\gamma_w\sqrt{s_w}}{\alpha}, \ \|\wh - \w_*\|_1 \leq \frac{12\gamma_w s_w}{\alpha}, \textrm{ and } \frac{\|\wh - \w_*\|_1}{\|\wh - \w_*\|_2} \leq 4\sqrt{s_w}
\]
provided (\ref{eqn:delta}) holds.
\end{thm}
When $\varsigma$ is small enough, the upper bound in Theorem~\ref{thm:new:approximate} is on the same order as that in Theorem~\ref{thm:main}. To be specific, we have the following corollary.
\begin{cor}
Assume $\|A^\top\w_*\|_2 = O(\sqrt{n})$, $\|\lbd_*\|_2=O(\sqrt{s_{\lambda}})$, $\alpha=O(\sqrt{n})$, $\beta=O(1)$, and $\varsigma = O(\sqrt{n \log n/m})$. When $m \geq O(s_\lambda \log  n)$, we can choose $\gamma_{\lambda}$ and $\gamma_{w}$ as in Corollary~\ref{cor:1} such that with a high probability
\[
\|\wh - \w_*\|_2 = O\left( \frac{\gamma_w\sqrt{s_w}}{\sqrt{n}} \right) = O\left(\sqrt{ \frac{s_w s_\lambda \log n}{m}}  \right) \textrm{ and } \frac{\|\wh - \w_*\|_1}{\|\wh - \w_*\|_2} \leq 4\sqrt{s_w}.
\]
\end{cor}

\subsection{More Results Under Stronger Assumptions} \label{sec:more:guarantee}
Under the stronger assumption that both columns and rows of $A$ have bounded $\ell_2$-norm, we have more ways to recover $\lbd_*$ and $\w_*$.
\subsubsection{Sparse Solutions}
Another approach for recovering $\lbd_*$ is still to solve the optimization problem in (\ref{eqn:cc-opt-1}) but with different settings of $\gamma_{\lambda}$ and $\gamma_w$.
\begin{thm}  \label{thm:main:2} Let  $(\wh, \lh)$ be the optimal solution to the problem in (\ref{eqn:cc-opt-1}). Define
\begin{equation} \label{eqn:cs:upper}
\zeta_{A^\top}(s) =  \inf\{ \delta: \|A^\top \z\|_2 \leq \delta \|\z\|_2, \ \forall \ \|\z\|_0 \leq s \}.
\end{equation}
Set
\begin{align*}
\gamma_{w} & \geq 2  \|\lbd_*\|_2 \sqrt{\frac{c}{m}\log \frac{4d}{\delta}}, \\
\gamma_{\lambda} & \geq  2   \|A^\top\w_*\|_2 \sqrt{\frac{c}{m}\log \frac{4 n}{\delta}}+\frac{6\gamma_w\sqrt{s_w}}{\alpha} \left(1 + 7 \zeta_{A^\top}(16s_w) \sqrt{\frac{c}{m} \left(\log \frac{4 n}{\delta}+ 16s_w \log \frac{9 d}{8 s_w}\right)}\right).
\end{align*}
With a probability at least $1-3\delta$, we have
\[
\|\lh - \lbd_*\|_2 \leq \frac{3\gamma_{\lambda} \sqrt{s_\lambda}}{\beta}, \ \|\lh - \lbd_*\|_1 \leq \frac{12 \gamma_{\lambda} s_\lambda}{\beta}, \textrm{ and }  \frac{\|\lh - \lbd_*\|_1}{\|\lh - \lbd_*\|_2} \leq  4 \sqrt{ s_\lambda}
\]
provided (\ref{eqn:delta}) holds.
\end{thm}
The conditions in Corollary~\ref{cor:1} cannot be reused to simplify Theorem~\ref{thm:main:2}, so we omit the simplification here.

Again, due to the symmetry between $\lbd$ and $\w$, we have the following theorem for recovering $\w_*$ by solving the optimization problem in (\ref{eqn:for:lambda})
\begin{thm}  \label{thm:main:2:second}  Let  $(\wh, \lh)$ be the optimal solution to the problem in (\ref{eqn:for:lambda}). Define
\[
\zeta_{A}(s) =  \inf\{ \delta: \|A \z\|_2 \leq \delta \|\z\|_2, \ \forall \ \|\z\|_0 \leq s \}.
\]
Set
\begin{align*}
\gamma_{\lambda} & \geq 2  \|\w_*\|_2 \sqrt{\frac{c}{m}\log \frac{4n}{\delta}}, \\
\gamma_{w} & \geq  2   \|A \lbd_*\|_2 \sqrt{\frac{c}{m}\log \frac{4 d}{\delta}}+\frac{6\gamma_{\lambda}\sqrt{s_{\lambda}}}{\beta} \left(1 + 7 \zeta_{A}(16s_{\lambda}) \sqrt{\frac{c}{m} \left(\log \frac{4 d}{\delta}+ 16s_{\lambda} \log \frac{9 n}{8 s_{\lambda}}\right)}\right).
\end{align*}
With a probability at least $1-3\delta$, we have
\[
\|\wh - \w_*\|_2  \leq \frac{3\gamma_{w} \sqrt{s_w}}{\alpha}, \ \|\wh - \w_*\|_1  \leq \frac{12 \gamma_{w} s_w}{\alpha}, \textrm{ and } \frac{\|\wh - \w_*\|_1}{\|\wh - \w_*\|_2} \leq  4 \sqrt{ s_w}
\]
provided (\ref{eqn:delta}) holds.
\end{thm}
When columns of $A$ are well-bounded, $\zeta_{A}(16s_{\lambda}) \leq O(\sqrt{s_{\lambda}})$. Then, the above theorem can be simplified as follows.
\begin{cor}
Assume $\|A \lbd_*\|_2= O(\sqrt{d})$, $\|\w_*\|_2=O(1)$, $\alpha=O(\sqrt{n})$, $\beta=O(1)$, $d \leq O(n)$ and $\zeta_{A}(16s_{\lambda}) \leq O(\sqrt{s_{\lambda}})$. When $m \geq O(s_\lambda \log  n)$, we can choose
\[
\gamma_{\lambda} = O\left(\sqrt{\frac{\log n}{m}}  \right) \textrm{ and }  \gamma_{w} =  O\left(\sqrt{\frac{d \log d}{m}}+ \sqrt{\frac{s_{\lambda} \log n}{m}} + \frac{s_{\lambda} \log n}{m} \sqrt{s_{\lambda}}\right)
\]
such that with a high probability
\[
\|\wh - \w_*\|_2 \leq O\left( \frac{\gamma_w\sqrt{s_w}}{\sqrt{n}} \right) \leq O\left(\sqrt{ \frac{s_w s_\lambda \log n}{m}}  \right) \textrm{ and } \frac{\|\wh - \w_*\|_1}{\|\wh - \w_*\|_2} \leq 4\sqrt{s_w}.
\]
\end{cor}
\subsubsection{Approximately Sparse Solutions}

When the optimal solutions to (\ref{eqn:cc-opt}) are allowed to be approximately sparse, $g(\cdot)$ could be certain smooth regularizer, such as a mixture of $\|\cdot\|_2^2$ and $\|\cdot\|_p^p$ for $1 < p < 2$. In the following, we provide a supporting theorem for this special case. we denote by $\w_*'$ and $\lbd_*'$ the optimal solutions to (\ref{eqn:cc-opt}). To quantify the approximate sparsity of $\w_*'$ and $\lbd_*'$, we assume there exist two sparse vectors $\w_*$ and $\lbd_*$, with $\|\w_*\|_0 = s_w$ and $\|\lbd_*\|_0 = s_{\lambda}$ such that
\begin{equation} \label{eqn:tau}
\|\w_* - \w_*'\|_2 \leq \tau \textrm{ and }\|\lbd_* - \lbd'_*\|_2 \leq \tau
\end{equation}
for some small constant $\tau>0$. Furthermore, we assume both $g$ and $h$ are $\mu$-smooth, i.e.,
\begin{align}
\|\nabla g(\w_1)-\nabla g(\w_2) \|_2 & \leq \mu \|\w_1-\w_2\|,\ \forall \w_1, \w_2, \label{eqn:smooth:g} \\
\|\nabla h(\lbd_1)-\nabla h(\lbd_2) \|_2 & \leq \mu \|\lbd_1-\lbd_2\|,\ \forall \lbd_1, \lbd_2. \label{eqn:smooth:h}
\end{align}
\begin{thm} \label{thm:approximate} Let  $(\wh, \lh)$ be the optimal solution to the problem in (\ref{eqn:cc-opt-1}).  Assume (\ref{eqn:tau}), (\ref{eqn:smooth:g}), and (\ref{eqn:smooth:h}) hold.  Suppose  $\w_*'$ and $\lbd_*'$ lie in the interior of $\Omega$ and $\Delta$, respectively. Set
\begin{align*}
\gamma_{\lambda} &\geq 2 \|A^\top\w_*\|_2 \sqrt{\frac{c}{m}\log \frac{4 n}{\delta}} + 2(1+\mu) \tau,\\
\gamma_{w} &\geq 2 \|\lbd_*\|_2 \sqrt{\frac{c}{m}\log \frac{4d}{\delta}} + \frac{6\gamma_\lambda\sqrt{s_\lambda}}{\beta} \left(1+7 \sqrt{\frac{c}{m}\left(\log \frac{4d}{\delta} +  16s_\lambda \log \frac{9 n}{8s_\lambda}\right)}\right)+ 2(1+\mu) \tau.
\end{align*}
With a probability at least $1-3\delta$, we have
\[
\|\wh - \w_*\|_2 \leq \frac{3\gamma_w\sqrt{s_w}}{\alpha}, \ \|\wh - \w_*\|_1 \leq \frac{12\gamma_w s_w}{\alpha}, \textrm{ and } \frac{\|\wh - \w_*\|_1}{\|\wh - \w_*\|_2} \leq 4\sqrt{s_w}
\]
provided (\ref{eqn:delta}) holds.
\end{thm}
Similarly, if $(1+\mu) \tau=O\left(\sqrt{ n \log n / m}  \right)$, the conclusion in Corollary~\ref{cor:1} also holds here.
\section{Analysis}
In this section, we provide main proofs of our theoretical results. The omitted proofs can be found in the appendix.
\subsection{Proof of Theorem~\ref{thm:main}}
To facilitate the analysis, we introduce a pseudo optimization problem
\[
\max\limits_{\lbd \in \Delta}  \; - h(\lbd) - \w_*^{\top}\Ah R^{\top}\lbd  - \gamma_{\lambda} \|\lbd\|_1
\]
whose optimal solution is denoted by $\lt$. In the following, we will first discuss how to bound the difference between $\lt$ and $\lbd_*$, and then bound the difference between $\wh$ and $\w_*$ in a similar way.

From the optimality of $\lt$ and $\lbd_*$, we derive the following lemma to bound their difference.
\begin{lem} \label{lem:lambda}
Denote
\begin{equation}  \label{eqn:rho:lambda}
 \rho_{\lambda}= \left\|(RR^\top -I)A^{\top}\w_* \right\|_{\infty}.
\end{equation}
By choosing $\gamma_{\lambda} \geq 2 \rho_{\lambda}$, we have
\[
\|\lt - \lbd_*\|_2 \leq \frac{3\gamma_\lambda\sqrt{s_\lambda}}{\beta}, \ \|\lt - \lbd_*\|_1 \leq \frac{12\gamma_\lambda s_\lambda}{\beta}, \textrm{ and } \frac{\|\lt - \lbd_*\|_1}{\|\lt - \lbd_*\|_2} \leq 4\sqrt{s_\lambda}.
\]
\end{lem}

Based on the property of the random matrix $R$ described in Property~\ref{thm:jl}, we have the following lemma to bound $\rho_{\lambda}$ in (\ref{eqn:rho:lambda}).
\begin{lem} \label{lem:rho:lambda}  With a probability at least $1 - \delta$, we have
\[
 \rho_{\lambda}= \left\|(RR^\top -I)A^{\top}\w_* \right\|_{\infty}\leq   \|A^\top \w_*\|_2 \sqrt{\frac{c}{m}\log \frac{4 n}{\delta}}
\]
provided (\ref{eqn:delta}) holds.
\end{lem}

Combining Lemma~\ref{lem:lambda} with Lemma~\ref{lem:rho:lambda}, we immediately obtain the following lemma.
\begin{lem} \label{thm:lambda:tilde}
Set
\[
\gamma_{\lambda} \geq 2 \|A^\top\w_*\|_2 \sqrt{\frac{c}{m}\log \frac{4 n}{\delta}}.
\]
With a probability at least $1-\delta$, we have
\[
\|\lt - \lbd_*\|_2 \leq \frac{3\gamma_\lambda\sqrt{s_\lambda}}{\beta}, \ \|\lt - \lbd_*\|_1 \leq \frac{12\gamma_\lambda s_\lambda}{\beta}, \textrm{ and } \frac{\|\lt - \lbd_*\|_1}{\|\lt - \lbd_*\|_2} \leq 4\sqrt{s_\lambda}
\]
provided (\ref{eqn:delta}) holds.
\end{lem}

We are now in a position to formulate the key lemmas that lead to Theorem~\ref{thm:main}. Similar to Lemma~\ref{lem:lambda}, we introduce the following lemma to characterize the relation between $\wh$ and $\w_*$.
\begin{lem} \label{lem:w}
Denote
\begin{equation} \label{eqn:rho:w}
 \rho_{w}= \left\| A ( I-RR^{\top})\lbd_* \right\|_\infty + \left\|A RR^\top (\lbd_*-\lt)  \right\|_{\infty}.
\end{equation}
By choosing $\gamma_{w} \geq 2 \rho_{w}$, we have
\[
\|\wh - \w_*\|_2 \leq \frac{3\gamma_w\sqrt{s_w}}{\alpha}, \ \|\wh - \w_*\|_1 \leq \frac{12\gamma_w s_w}{\alpha}, \textrm{ and } \frac{\|\wh - \w_*\|_1}{\|\wh - \w_*\|_2} \leq 4\sqrt{s_w}.
\]
\end{lem}

The last step of the proof is to derive an upper bound for $\rho_{w}$ based on Property~\ref{thm:jl} and Lemma~\ref{thm:lambda:tilde}.
\begin{lem} \label{lem:rho:w} Assume the conclusion in Lemma~\ref{thm:lambda:tilde} happens.  With a probability at least $1 - 2 \delta$, we have
\[
 \rho_{w}\leq \|\lbd_*\|_2 \sqrt{\frac{c}{m}\log \frac{4d}{\delta}} + \frac{3\gamma_\lambda\sqrt{s_\lambda}}{\beta} \left(1+7 \sqrt{\frac{c}{m}\left(\log \frac{4d}{\delta} +  16s_\lambda \log \frac{9 n}{8s_\lambda}\right)}\right)
\]
provided (\ref{eqn:delta}) holds.
\end{lem}

\subsection{Proof of Lemma~\ref{lem:lambda}} \label{sec:lem:lambda}
\paragraph{Notations} For a vector $\x \in \R^d$ and a set $\D \subseteq [d]$, we denote by  $\x_{\D}$  the vector which coincides with $\x$ on $\D$ and has zero coordinates outside $\D$.

Let $\Ol$ include the subset of non-zeros entries in $\lbd_*$ and  $\Obl=[n]\setminus \Ol$. Define
\[
\begin{split}
\L(\lbd) & = -h(\lbd) + \min\limits_{\w \in \Omega} g(\w)-  \w^{\top}A\lbd,\\
\Lt(\lbd) & =  - h(\lbd) - \w_*^{\top}\Ah R^{\top}\lbd  - \gamma_{\lambda} \|\lbd\|_1.\\
\end{split}
\]

Let $\v \in \partial \|\lbd_*\|_1$ be any subgradient of  $\|\cdot\|_1$ at $\lbd_*$. Then, we have
\[
\u=-\nabla h(\lbd_*)-  RR^{\top}A^{\top}\w_*- \gamma_{\lambda} \v  \in   \partial \Lt(\lbd_*).~\footnote{In the case that $h(\cdot)$ is non-smooth, $\nabla h(\lbd_*)$ refers to a subgradient of $h(\cdot)$ at $\lbd_*$. In particular, we choose the subgradient that satisfies (\ref{eqn:lem1:4}).}
\]
Using the fact that $\lt$ maximizes $\Lt(\cdot)$ over the domain $\Delta$ and $h(\cdot)$ is $\beta$-strongly convex, we have
\begin{equation} \label{eqn:lem1:1}
\begin{split}
0& \geq \Lt(\lbd_*)- \Lt(\lt)  \geq  \langle-(\lt - \lbd_*), \u \rangle + \frac{\beta}{2}\|\lbd_* - \lt\|_2^2 \\
=&\left\langle\lt - \lbd_*, \nabla h(\lbd_*)+  RR^{\top}A^{\top}\w_* + \gamma_{\lambda} \v \right\rangle + \frac{\beta}{2}\|\lbd_* - \lt\|_2^2 .\\
\end{split}
\end{equation}
By setting $v_i=  \sgn(\widetilde{\lambda}_i)$, $\forall  i \in \Obl$, we have $\langle \lt_{\Obl} , \v_{\Obl} \rangle  = \|\lt_{\Obl}\|_1$. As a result,
\begin{equation} \label{eqn:lem1:2}
\langle \lt - \lbd_*, \v \rangle = \langle \lt_{\Obl} , \v_{\Obl} \rangle+ \langle \lt_{\Ol}- \lbd_* , \v_{\Ol} \rangle \geq \|\lt_{\Obl}\|_1 - \|\lt_{\Ol} - \lbd_*\|_1.
\end{equation}
Combining (\ref{eqn:lem1:1}) with (\ref{eqn:lem1:2}), we have
\begin{equation} \label{eqn:lem1:3}
\left \langle \lt - \lbd_*, \nabla h(\lbd_*) + RR^{\top}A^{\top}\w_* \right\rangle+\frac{\beta}{2}\|\lbd_* - \lt\|_2^2 + \gamma_{\lambda}\|\lt_{\Obl}\|_1 \leq  \gamma_{\lambda}\|\lt_{\Ol} - \lbd_*\|_1 .
\end{equation}

From the fact that $\lbd_*$ maximizes $\L(\cdot)$ over the domain $\Delta$, we have
\begin{equation} \label{eqn:lem1:4}
\langle \nabla \L(\lbd_*), \lbd- \lbd_* \rangle = \langle -\nabla h(\lbd_*) - A^\top \w_*, \lbd- \lbd_* \rangle \leq 0, \ \forall \lbd \in \Delta.
\end{equation}
Then,
\begin{equation} \label{eqn:lem1:5}
\begin{split}
& \left \langle \lt - \lbd_*, \nabla h(\lbd_*) + RR^\top A^{\top}\w_*\right\rangle \\
= & \left\langle \lt - \lbd_*, \nabla h(\lbd_*) + A^{\top}\w_* \right\rangle +  \left \langle \lt - \lbd_*, (RR^{\top} -I)A^{\top}\w_* \right\rangle\\
\overset{\text{(\ref{eqn:lem1:4})}}{\geq}  & -\|\lt - \lbd_*\|_1 \left\|(RR^{\top} -I)A^{\top}\w_* \right\|_{\infty} \\
\overset{\text{(\ref{eqn:rho:lambda})}}{=}  &  -\rho_{\lambda} \|\lt - \lbd_*\|_1=-\rho_{\lambda} \left( \|\lt_{\Obl}\|_1  + \|\lt_{\Ol} - \lbd_*\|_1\right).
\end{split}
\end{equation}

From  (\ref{eqn:lem1:3}) and  (\ref{eqn:lem1:5}), we have
\[
\frac{\beta}{2}\|\lt - \lbd_*\|_2^2 + (\gamma_{\lambda} - \rho_{\lambda})\|\lt_{\Obl}\|_1 \leq (\gamma_{\lambda} + \rho_{\lambda})\|\lt_{\Ol} - \lbd_*\|_1.
\]
Since $\gamma_{\lambda} \geq 2 \rho_{\lambda}$, we have
\[
\frac{\beta}{2}\|\lt - \lbd_*\|_2^2 + \frac{\gamma_{\lambda}}{2}\|\lt_{\Obl}\|_1 \leq \frac{3\gamma_{\lambda}}{2}\|\lt_{\Ol} - \lbd_*\|_1.
\]
And thus,
\[
\begin{split}
\frac{\beta}{2}\|\lt - \lbd_*\|_2^2 \leq \frac{3\gamma_{\lambda}}{2}\|\lt_{\Ol} - \lbd_*\|_1 \leq \frac{3\gamma_{\lambda} \sqrt{s_\lambda}}{2}\|\lt_{\Ol} - \lbd_*\|_2 \Rightarrow  \|\lt - \lbd_*\|_2 \leq \frac{3\gamma_{\lambda} \sqrt{s_\lambda}}{\beta}, \\
\frac{\beta }{2 s_\lambda}\|\lt_{\Ol} - \lbd_*\|_1^2   \leq \frac{\beta}{2}\|\lt - \lbd_*\|_2^2 \leq \frac{3\gamma_{\lambda}}{2}\|\lt_{\Ol} - \lbd_*\|_1 \Rightarrow \|\lt_{\Ol} - \lbd_*\|_1 \leq  \frac{3\gamma_{\lambda} s_\lambda}{\beta}, \\
\frac{\gamma_{\lambda}}{2}\|\lt_{\Obl}\|_1 \leq \frac{3\gamma_{\lambda}}{2}\|\lt_{\Ol} - \lbd_*\|_1 \Rightarrow  \|\lt_{\Obl}\|_1 \leq  3\|\lt_{\Ol} - \lbd_*\|_1 \Rightarrow \|\lt - \lbd_*\|_1 \leq \frac{12\gamma_{\lambda} s_\lambda}{\beta},\\
\frac{\|\lt - \lbd_*\|_1}{\|\lt - \lbd_*\|_2}  =\frac{\|\lt_{\Ol} - \lbd_*\|_1 + \|\lt_{\Obl}\|_1}{\|\lt - \lbd_*\|_2} \leq \frac{4 \|\lt_{\Ol} - \lbd_*\|_1 }{\|\lt - \lbd_*\|_2} \leq \frac{4 \sqrt{ s_\lambda}\|\lt_{\Ol} - \lbd_*\|_2}{\|\lt - \lbd_*\|_2} \leq 4 \sqrt{ s_\lambda}.
\end{split}
\]
\subsection{Proof of Lemma~\ref{lem:rho:lambda}} \label{sec:first:con}
We first introduce one lemma that is central to our analysis. From the property that $R$ preserves the $\ell_2$-norm, it is easy to verify that it also preserves the inner product~\citep{ML06:Arriaga}. Specifically, we have the following lemma.
\begin{lem} \label{lem:inner} Assume $R$ satisfies Property~\ref{thm:jl}.  For any two fixed vectors $\u \in \R^n$ and $\v\in \R^n$, with a probability at least $1-\delta$, we have
\[
   \left|\u^{\top} RR^\top \v - \u^{\top}\v\right| \leq \|\u\|_2\|\v\|_2   \sqrt{\frac{c}{m}\log \frac{4}{\delta}}.
\]
provided (\ref{eqn:delta}) holds.
\end{lem}

Let $\e_j$ be the  $j$-th standard basis vector of $\R^n$. From Lemma~\ref{lem:inner},  we have with a probability at least $1-\delta$,
\[
\left|\left[ (RR^\top- I)A^{\top}\w_* \right]_j\right|=\left|\e_j^\top (RR^{\top} - I) A^{\top}\w_*\right| \leq \|A^{\top}\w_*\|_2 \sqrt{\frac{c}{m}\log \frac{4}{\delta}}
\]
for each $j \in [n]$. We complete the proof by taking the union bound over all $j \in [n]$.
\subsection{Proof of Lemma~\ref{lem:w}} \label{sec:lem:w}
Let $\Ow$ include the subset of non-zeros entries in $\w_*$ and $\Obw=[d] \setminus \Ow$.
Define
\[
\begin{split}
\G(\w) & =  g(\w) + \max\limits_{\lbd \in \Delta} - h(\lbd) - \w^{\top}A\lbd,\\
\Gh(\w)& =g(\w) + \gamma_{w}\|\w\|_1 +\max_{\lbd \in \Delta}- h(\lbd) - \w^{\top}\Ah R^{\top}\lbd  - \gamma_{\lambda} \|\lbd\|_1.
\end{split}
\]

Let $\v \in \partial \|\w_*\|_1$ be any subgradient of  $ \|\cdot\|_1$ at $\w_*$. Then, we have
\[
\u = \nabla g(\w_*)- ARR^{\top}\lt + \gamma_w  \v \in  \partial \Gh(\w_*).~\footnote{In the case that $g(\cdot)$ is non-smooth, $\nabla g(\w_*)$ refers to a subgradient of $g(\cdot)$ at $\w_*$. In particular, we choose the subgradient that satisfies (\ref{eqn:opt:new}). }
\]
Using the fact that $\wh$ minimizes $\Gh(\cdot)$ over the domain $\Omega$ and $g(\cdot)$ is $\alpha$-strongly convex, we have
\begin{equation} \label{eqn:1:new}
\begin{split}
0& \geq \Gh(\wh)- \Gh(\w_*)  \geq  \langle \wh-\w_*,  \u \rangle + \frac{\alpha}{2}\|\wh-\w_*\|_2^2 \\
=&\left \langle \wh-\w_*, \nabla g(\w_*)- ARR^{\top} \lt + \gamma_w  \v \right\rangle + \frac{\alpha}{2}\|\wh-\w_*\|_2^2.\\
\end{split}
\end{equation}
By setting $v_i= \sgn(\widehat{w}_i)$, $\forall i \in \Obw$, we have $\langle \wh_{\Obw} , \v_{\Obw} \rangle  = \|\wh_{\Obw}\|_1$.  As a result,
\begin{equation} \label{eqn:2:new}
\langle \wh-\w_* , \v \rangle = \langle \wh_{\Obw}, \v_{\Obw}\rangle+ \langle \wh_{\Ow} - \w_* , \v_{\Ow} \rangle \geq \|\wh_{\Obw}\|_1 - \|\wh_{\Ow} - \w_*\|_1.
\end{equation}
Combining (\ref{eqn:1:new}) with (\ref{eqn:2:new}), we have
\begin{equation} \label{eqn:3:new}
\left \langle \wh-\w_*, \nabla g(\w_*)- ARR^{\top}\lt \right\rangle+\frac{\alpha}{2}\|\wh-\w_*\|_2^2 + \gamma_w  \|\wh_{\Obw}\|_1 \leq  \gamma_w  \|\wh_{\Ow} - \w_*\|_1 .
\end{equation}

From the fact that $\w_*$ minimizes $\G(\cdot)$ over the domain $\Omega$, we have
\begin{equation} \label{eqn:opt:new}
\langle \nabla \G(\w_*), \w- \w_* \rangle = \langle \nabla g(\w_*) - A \lbd_*, \w- \w_* \rangle \geq 0, \ \forall \w \in \Omega.
\end{equation}
Then,
\begin{equation} \label{eqn:4:new}
\begin{split}
& \left \langle \wh-\w_*, \nabla g(\w_*)- ARR^{\top} \lt \right\rangle \\
= &  \left\langle \wh-\w_*, \nabla g(\w_*)-  A \lbd_* \right\rangle + \left \langle \wh-\w_*, A (I- RR^{\top})\lbd_* \right\rangle +  \left \langle \wh-\w_*, A RR^\top (\lbd_*-\lt) \right\rangle\\
\overset{\text{(\ref{eqn:opt:new})}}{\geq}  & -\|\wh-\w_*\|_1  \left(\left\|A (I- RR^{\top})\lbd_* \right\|_{\infty} + \left\|A RR^\top (\lbd_*-\lt)  \right\|_{\infty}\right)\\
\overset{\text{(\ref{eqn:rho:w})}}{=}  &-\rho_{w} \|\wh-\w_*\|_1 =-\rho_{w} \left(\|\wh_{\Obw}\|_1 + \|\wh_{\Ow} - \w_*\|_1\right).
\end{split}
\end{equation}

From  (\ref{eqn:3:new}) and  (\ref{eqn:4:new}), we have
\[
\frac{\alpha}{2}\|\wh-\w_*\|_2^2  + (\gamma_{w} - \rho_{w}) \|\wh_{\Obw}\|_1 \leq (\gamma_{w} + \rho_{w})\|\wh_{\Ow} - \w_*\|_1.
\]
Since $\gamma_{w} \geq 2 \rho_{w}$, we have
\[
\frac{\alpha}{2}\| \wh-\w_* \|_2^2 + \frac{\gamma_{w}}{2} \|\wh_{\Obw}\|_1 \leq \frac{3\gamma_{w}}{2} \|\wh_{\Ow} - \w_*\|_1.
\]
And thus,
\[
\begin{split}
\frac{\alpha}{2}\|\wh-\w_*\|_2^2 \leq \frac{3\gamma_{w}}{2}\|\wh_{\Ow} - \w_*\|_1 \leq \frac{3\gamma_{w} \sqrt{s_w}}{2}\|\wh_{\Ow} - \w_*\|_2 \Rightarrow  \|\wh-\w_*\|_2 \leq \frac{3\gamma_{w} \sqrt{s_w}}{\alpha} \\
\frac{\alpha }{2 s_w}\|\wh_{\Ow} - \w_*\|_1^2   \leq \frac{\alpha}{2}\|\wh-\w_*\|_2^2 \leq \frac{3\gamma_{w}}{2}\|\wh_{\Ow} - \w_*\|_1  \Rightarrow \|\wh_{\Ow} - \w_*\|_1 \leq  \frac{3\gamma_{w} s_w}{\alpha} \\
\frac{\gamma_{w}}{2} \|\wh_{\Obw}\|_1  \leq \frac{3\gamma_{w}}{2} \|\wh_{\Ow} - \w_*\|_1 \Rightarrow \|\wh_{\Obw}\|_1 \leq  3\|\wh_{\Ow} - \w_*\|_1 \Rightarrow \|\wh-\w_*\|_1 \leq \frac{12\gamma_{w} s_w}{\alpha}\\
\frac{\|\wh - \w_*\|_1}{\|\wh - \w_*\|_2} = \frac{\|\wh_{\Ow} - \w_*\|_1 + \|\wh_{\Obw}\|_1}{\|\wh - \w_*\|_2} \leq \frac{4\|\wh_{\Ow} - \w_*\|_1}{\|\wh- \w_*\|_2} \leq \frac{4\sqrt{s_w}\|\wh_{\Ow} - \w_*\|_2}{\|\wh- \w_*\|_2} \leq  4\sqrt{s_w}.
\end{split}
\]
\subsection{Proof of Lemma~\ref{lem:rho:w}} \label{sec:lem:rho:w}
We first upper bound $\rho_w$ as
\[
 \rho_{w} \leq  \underbrace{\left\| A ( I-RR^{\top})\lbd_* \right\|_\infty}_{:=U_1} + \underbrace{\left\|A (\lbd_*-\lt)  \right\|_{\infty}}_{:=U_2} + \underbrace{\left\|A (RR^\top -I)(\lbd_*-\lt)  \right\|_{\infty}}_{:=U_3}.
\]

\paragraph{Bounding $U_1$}  From Lemma~\ref{lem:inner},  we have with a probability at least $1-\delta$,
\[
\begin{split}
& \left|\left[A ( I-RR^{\top})\lbd_* \right]_i\right|=\left|A_{i*} ( I-RR^{\top})\lbd_* \right| \\
\leq & \max_{i\in[d]} \|A_{i*}\|_2 \|\lbd_*\|_2 \sqrt{\frac{c}{m}\log \frac{4}{\delta}} \overset{\text{(\ref{eqn:upper:A1})}}{\leq} \|\lbd_*\|_2 \sqrt{\frac{c}{m}\log \frac{4}{\delta}}
 \end{split}
\]
for each $i \in [d]$. Taking the union bound over all $i \in [d]$,  we have with a probability at least $1-\delta$,
\[
\left\| A ( I-RR^{\top})\lbd_* \right\|_\infty \leq \|\lbd_*\|_2 \sqrt{\frac{c}{m}\log \frac{4d}{\delta}}.
\]

\paragraph{Bounding $U_2$} From our assumption, we have
\[
\left\|A (\lbd_*-\lt)  \right\|_{\infty} \leq \max_{i\in[d]} \|A_{i*}\|_2 \|\lbd_*-\lt\|_2 \overset{\text{(\ref{eqn:upper:A1})}}{\leq} \|\lbd_*-\lt\|_2.
\]

\paragraph{Bounding $U_3$} Notice that the arguments for bounding $U_1$ cannot be used to upper bound $U_3$, that is because $\lbd_*-\lt$ is a random variable that depends on $R$ and thus we cannot apply Lemma~\ref{lem:inner} directly. To overcome this challenge, we will exploit the fact that $\lbd_*-\lt$ is approximately sparse to decouple the dependence.  Define
\[
\begin{split}
\K_{n,16 s_\lambda} =  \left\{\x \in \R^n: \|\x\|_2 \leq 1, \|\x\|_1 \leq 4\sqrt{s_\lambda} \right\}.
\end{split}
\]
When the conclusion in Lemma~\ref{thm:lambda:tilde} happens, we have
\begin{equation} \label{eqn:sparse:belong}
\frac{\lt - \lbd_*}{\|\lt - \lbd_*\|_2} \in \K_{n,16 s_\lambda}
\end{equation}
and thus
\[
\begin{split}
&U_3 = \|\lbd_*-\lt\|_2  \left\|A (RR^\top -I ) \frac{\lbd_*-\lt}{\|\lbd_*-\lt\|_2}  \right\|_{\infty} \\
\overset{\text{(\ref{eqn:sparse:belong})}}{\leq} & \|\lbd_*-\lt\|_2  \underbrace{\sup_{\z \in \K_{n,16 s_\lambda}} \left\|A (RR^\top -I) \z \right\|_{\infty}}_{:=U_4}.
\end{split}
\]
Then, we will utilize techniques of covering number to provide an upper bound for $U_4$.
\begin{lem} \label{lem:conver} With a probability at least $1-\delta$, we have
\[
\sup_{\z \in \K_{n,16 s_\lambda}} \left\|A (RR^\top -I) \z \right\|_{\infty} \leq 2(2+\sqrt{2})  \sqrt{\frac{c}{m}\left(\log \frac{4d}{\delta} +  16s_\lambda \log \frac{9 n}{8s_\lambda}\right)}.
\]
\end{lem}

Putting everything together, we have
\[
\begin{split}
&\rho_{w} \\
\leq & \|\lbd_*\|_2 \sqrt{\frac{c}{m}\log \frac{4d}{\delta}} \\
&+  \|\lbd_*-\lt\|_2 \left(1+2(2+\sqrt{2})  \sqrt{\frac{c}{m}\left(\log \frac{4d}{\delta} +  16s_\lambda \log \frac{9 n}{8s_\lambda}\right)}\right) \\
\leq & \|\lbd_*\|_2 \sqrt{\frac{c}{m}\log \frac{4d}{\delta}} + \frac{3\gamma_\lambda\sqrt{s_\lambda}}{\beta} \left(1+7 \sqrt{\frac{c}{m}\left(\log \frac{4d}{\delta} +  16s_\lambda \log \frac{9 n}{8s_\lambda}\right)}\right).
\end{split}
\]
\subsection{Proof of Lemma~\ref{lem:inner}}
First, we assume $\|\u\|_2=\|\v\|_2=1$. Following the proof of Corollary 2 in \citep{ML06:Arriaga}, we apply Property~\ref{thm:jl} to vectors $\u+\v$ and $\u-\v$. Then, with a probability at least $1 - 4\exp(- m \varepsilon^2 /c)$, we have
\begin{align}
(1 - \varepsilon)\|\u+\v\|_2^2 \leq \|R^\top (\u+\v)\|_2^2 \leq (1 + \varepsilon)\|\u+\v\|_2^2, \label{eqn:inner:1} \\
(1 - \varepsilon)\|\u-\v\|_2^2 \leq \|R^\top (\u-\v)\|_2^2 \leq (1 + \varepsilon)\|\u-\v\|_2^2,\label{eqn:inner:2}
\end{align}
provided $\epsilon \leq 1/2$. From (\ref{eqn:inner:1}) and (\ref{eqn:inner:2}), it is straightforward to show that
\[
\left|\u^{\top} RR^\top  \v - \u^{\top}\v \right| \leq  \varepsilon.
\]
Thus,  with a probability at least $1 - \delta$, we have
\[
\left|\u^{\top} RR^\top \v - \u^{\top}\v\right| \leq  \sqrt{\frac{c}{m}\log \frac{4}{\delta}}
\]
provided (\ref{eqn:delta}) holds.

We complete the proof by noticing
\[
 \left|\u^{\top}RR^{\top} \v - \u^{\top}\v \right | =  \|\u\|_2 \|\v\|_2 \left|\frac{\u^{\top}}{\|\u\|_2}RR^{\top} \frac{\v}{\|\v\|_2} - \frac{\u^{\top}\v}{\|\u\|_2 \|\v\|_2} \right|.
\]
\subsection{Proof of Lemma~\ref{lem:conver}}
First, we define
\[
\S_{n,16s_\lambda}=  \left\{\x \in \R^n: \|\x\|_2 \leq 1, \|\x\|_0 \leq 16 s_\lambda \right\} .
\]
Using Lemma 3.1 from~\citep{OneBit:Plan:LP}, we have $\K_{n,16 s_\lambda}  \subset 2 \conv(\S_{n,16s_\lambda})$ and therefore
\begin{equation} \label{eqn:covering:1}
U_4 \leq 2 \sup_{\z \in \conv(\S_{n,16s_\lambda})} \left\|A (RR^\top -I) \z \right\|_{\infty} =  2 \underbrace{\sup_{\z \in \S_{n,16s_\lambda}} \left\|A (RR^\top -I) \z \right\|_{\infty}}_{:=\theta}
\end{equation}
where the last equality follows from the fact that the maximum of a convex function over a convex set generally occurs at some extreme point of the set~\citep{Convex:analysis}.

Let $\S_{n,s}(\epsilon)$ be a proper $\epsilon$-net for $\S_{n,s}$ with the smallest cardinality, and $|\S_{n,s}(\epsilon)|$ be the covering number for $\S_{n,s}$. We have the following lemma for bounding $|\S_{n,s}(\epsilon)|$.
\begin{lem} \cite[Lemma 3.3]{OneBit:Plan:LP} \label{lemma:covering-number} For $\epsilon \in (0,1)$ and $s \leq n$, we have
\[
    \log |\S_{n,s}(\epsilon)| \leq s \log\left( \frac{9 n}{\epsilon s}\right).
\]
\end{lem}
Let  $\S_{n,16s_\lambda}(\epsilon)$ be a $\epsilon$-net of $\S_{n,16s_\lambda}$ with smallest cardinality.
With the help of $\S_{n,16s_\lambda}(\epsilon)$, we define a discretized version of $\theta$ in (\ref{eqn:covering:1}) as
\[
\theta(\epsilon) = \sup\left\{\left\|A (RR^\top -I) \z\right\|_{\infty}: \z \in \S_{n,16s_\lambda}(\epsilon)\right\}.
\]
The following lemma relates $\theta$ with $\theta(\epsilon)$.
\begin{lem} \cite[Lemma 9.2]{Oracle_inequality}\label{lemma:discrete-bound}
For $\epsilon \in (0, 1/\sqrt{2})$, we have
\[
\theta \leq \frac{\theta(\epsilon)}{1-\sqrt{2} \epsilon}.
\]
\end{lem}
By choosing $\epsilon=1/2$, we have $\theta \leq (2+\sqrt{2}) \theta(1/2)$. Combining with (\ref{eqn:covering:1}), we obtain
\[
U_4 \leq 2(2+\sqrt{2}) \underbrace{\sup\left\{\left\|A (RR^\top -I) \z\right\|_{\infty}: \z \in \S_{n,16s_\lambda}(1/2)\right\}}_{\theta(1/2)}
\]
Furthermore, Lemma~\ref{lemma:covering-number} implies
\[
    \log |\S_{n,16s_\lambda}(1/2)| \leq 16s_\lambda \log\left( \frac{9 n}{8s_\lambda}\right).
\]

We proceed by providing an upper bound for $\theta(1/2)$. Following the arguments for bounding $U_1$ in the proof of Lemma~\ref{lem:rho:w}, we have with a probability at least $1-\delta$,
\[
\left\| A \left( RR^{\top}-I\right) \z  \right\|_\infty  \leq \sqrt{\frac{c}{m}\log \frac{4d}{\delta}}
\]
for each $\z \in \S_{n,16s_\lambda}(1/2)$. We complete the proof by taking the union bound over all $\z \in \S_{n,16s_\lambda}(1/2)$.
\subsection{Proof of Theorem~\ref{thm:new:approximate}}
The proof is almost identical to that of Theorem~\ref{thm:main}. We just need to replace Lemmas~\ref{lem:lambda} and \ref{lem:w} with the following ones.

\begin{lem} \label{lem:lambda:new:app}
Denote
\begin{equation}  \label{eqn:rho:lambda:new:app}
 \rho_{\lambda}= \left\|(RR^\top -I)A^{\top}\w_* \right\|_{\infty}+ \varsigma.
\end{equation}
By choosing $\gamma_{\lambda} \geq 2 \rho_{\lambda}$, we have
\[
\|\lt - \lbd_*\|_2 \leq \frac{3\gamma_\lambda\sqrt{s_\lambda}}{\beta}, \ \|\lt - \lbd_*\|_1 \leq \frac{12\gamma_\lambda s_\lambda}{\beta}, \textrm{ and } \frac{\|\lt - \lbd_*\|_1}{\|\lt - \lbd_*\|_2} \leq 4\sqrt{s_\lambda}.
\]
\end{lem}
\begin{lem} \label{lem:w:new:app}
Denote
\begin{equation} \label{eqn:rho:w:new:app}
 \rho_{w}= \left\| A \left( I-RR^{\top}\right)\lbd_* \right\|_\infty + \left\|A RR^\top (\lbd_*-\lt)  \right\|_{\infty}+\varsigma.
\end{equation}
By choosing $\gamma_{w} \geq 2 \rho_{w}$, we have
\[
\|\wh - \w_*\|_2 \leq \frac{3\gamma_w\sqrt{s_w}}{\alpha}, \ \|\wh - \w_*\|_1 \leq \frac{12\gamma_w s_w}{\alpha}, \textrm{ and } \frac{\|\wh - \w_*\|_1}{\|\wh - \w_*\|_2} \leq 4\sqrt{s_w}.
\]
\end{lem}

\section{Conclusion and Future Work}
In this paper, a randomized algorithm is proposed to solve the convex-concave optimization problem in (\ref{eqn:cc-opt}). Compared to previous studies, a distinctive feature of the proposed algorithm is that  $\ell_1$-norm regularization is introduced to control the damage cased by random projection. Under mild assumptions about the optimization problem, we demonstrate that it is able to accurately recover the optimal solutions to (\ref{eqn:cc-opt}) provided they are sparse or approximately sparse.

From the current analysis, we need to solve two different problems if our goal is to recover both $\w_*$ and $\lbd_*$ accurately. It is unclear whether this is an artifact of the proof technique or actually unavoidable. We will investigate this issue in the future. Since the proposed algorithm is designed for the case that the optimal solutions are (approximately) sparse, it is practically important to develop a pre-precessing procedure that can estimate the sparsity of solutions before applying our algorithm. We plan to utilize random sampling to address this problem. Last but not least, we will investigate the empirical performance of the proposed algorithm.




\vskip 0.2in
\bibliography{E:/MyPaper/ref}

\appendix
\section{Proof of Theorem~\ref{thm:2}}
The analysis here is similar to that for Lemma~\ref{lem:lambda}.  Recall that in the proof of Theorem~\ref{thm:main}, we have proved that
\begin{equation} \label{eqn:inter:gamma:lambda}
\gamma_{\lambda} \geq 2 \|A^\top\w_*\|_2 \sqrt{\frac{c}{m}\log \frac{4 n}{\delta}} \geq 2 \left\|(RR^\top -I)A^{\top}\w_* \right\|_{\infty}
\end{equation}
holds with a probability at least $1-\delta$.

Define
\[
\Lh(\lbd) =  - h(\lbd) - \wh^{\top}\Ah R^{\top}\lbd  - \gamma_{\lambda} \|\lbd\|_1.\\
\]
Using the fact that $\lh$ maximizes $\Lh(\cdot)$ over the domain $\Delta$ and $h(\cdot)$ is $\beta$-strongly convex, we have
\begin{equation} \label{eqn:lem2:3}
\left \langle \lh - \lbd_*, \nabla h(\lbd_*) + RR^{\top}A^{\top}\wh \right\rangle+\frac{\beta}{2}\|\lbd_* - \lh\|_2^2 + \gamma_{\lambda}\|\lh_{\Obl}\|_1 \leq  \gamma_{\lambda}\|\lh_{\Ol} - \lbd_*\|_1 .
\end{equation}

On the other hand, we have
\begin{equation} \label{eqn:lem2:5}
\begin{split}
& \left \langle \lh - \lbd_*, \nabla h(\lbd_*) + RR^\top A^{\top}\wh\right\rangle \\
= & \langle \lh - \lbd_*, \nabla h(\lbd_*) + A^{\top}\w_* \rangle +  \left \langle \lh - \lbd_*, (RR^{\top} -I)A^{\top}\w_* \right\rangle+  \left \langle \lh - \lbd_*, RR^{\top} A^{\top}(\wh-\w_*) \right\rangle\\
\overset{\text{(\ref{eqn:lem1:4})}}{\geq}  & -\|\lh - \lbd_*\|_1  \left\|(RR^{\top} -I)A^{\top}\w_* \right\|_{\infty} -\|\lh - \lbd_*\|_2 \left\|RR^{\top} A^{\top}(\wh-\w_*)\right\|_{2}\\
\overset{\text{(\ref{eqn:inter:gamma:lambda})}}{\geq}  &  - \frac{\gamma_{\lambda}}{2}\|\lh - \lbd_*\|_1 -\|\lh - \lbd_*\|_2 \left\|RR^{\top} A^{\top}(\wh-\w_*)\right\|_2.
\end{split}
\end{equation}

From  (\ref{eqn:lem2:3}) and  (\ref{eqn:lem2:5}), we have
\[
\begin{split}
&\frac{\beta}{2}\|\lbd_* - \lh\|_2^2+ \frac{\gamma_{\lambda}}{2}  \|\lh_{\Obl}\|_1 \\
\leq  & \frac{3\gamma_{\lambda}}{2}\|\lt_{\Ol} - \lbd_*\|_1 +\|\lh - \lbd_*\|_2 \left\|RR^{\top} A^{\top}(\wh-\w_*)\right\|_2 \\
\leq & \frac{3\gamma_{\lambda} \sqrt{s_\lambda}}{2}\|\lt_{\Ol} - \lbd_*\|_2 +\|\lh - \lbd_*\|_2 \left\|RR^{\top} A^{\top}(\wh-\w_*)\right\|_2 \\
\leq & \|\lh - \lbd_*\|_2 \left(\frac{3\gamma_{\lambda} \sqrt{s_\lambda}}{2}+ \left\|RR^{\top} A^{\top}(\wh-\w_*)\right\|_2\right)
\end{split}
\]
which implies
\[
\begin{split}
&\|\lbd_* - \lh\|_2 \\
\leq &  \frac{2}{\beta} \left(\frac{3\gamma_{\lambda} \sqrt{s_\lambda}}{2}+ \left\|RR^{\top} A^{\top}(\wh-\w_*)\right\|_2\right) \\
\leq &  \frac{2}{\beta} \left(\frac{3\gamma_{\lambda} \sqrt{s_\lambda}}{2}+ \|A^{\top}(\wh-\w_*)\|_2+\left\|(RR^{\top}-I) A^{\top}(\wh-\w_*)\right\|_2\right) \\
\leq &  \frac{2}{\beta} \left(\frac{3\gamma_{\lambda} \sqrt{s_\lambda}}{2}+ \left( 1+ \|RR^{\top}-I\|_2\right)\|A^{\top}(\wh-\w_*)\|_2\right) .
\end{split}
\]
\section{Proof of Lemma~\ref{lem:lambda:new:app}}
From the assumption, we have
\[
\begin{split}
& \left \langle \lt - \lbd_*, \nabla h(\lbd_*) + RR^\top A^{\top}\w_*\right\rangle \\
= & \left\langle \lt - \lbd_*, \nabla h(\lbd_*) + A^{\top}\w_* \right\rangle +  \left \langle \lt - \lbd_*, (RR^{\top} -I)A^{\top}\w_* \right\rangle\\
\overset{\text{(\ref{eqn:app:second:2})}}{\geq}  & -\|\lt - \lbd_*\|_1 \left(\left\|(RR^{\top} -I)A^{\top}\w_*\right\|_{\infty}+ \varsigma \right)\\
\overset{\text{(\ref{eqn:rho:lambda:new:app})}}{=}  &  -\rho_{\lambda} \|\lt - \lbd_*\|_1=-\rho_{\lambda} \left( \|\lt_{\Obl}\|_1  + \|\lt_{\Ol} - \lbd_*\|_1\right).
\end{split}
\]
Substituting the above inequality into (\ref{eqn:lem1:3}), and the rest proof is identical to that of Lemma~\ref{lem:lambda}.
\section{Proof of Lemma~\ref{lem:w:new:app}}
Similarly, we have
\[
\begin{split}
& \left \langle \wh-\w_*, \nabla g(\w_*)- ARR^{\top} \lt \right\rangle \\
= &  \left\langle \wh-\w_*, \nabla g(\w_*)-  A \lbd_* \right\rangle + \left \langle \wh-\w_*, A (I- RR^{\top})\lbd_* \right\rangle +  \left \langle \wh-\w_*, A RR^\top (\lbd_*-\lt) \right\rangle\\
\overset{\text{(\ref{eqn:app:second:1})}}{\geq}  & -\|\wh-\w_*\|_1  \left( \left\|A (I- RR^{\top})\lbd_* \right\|_{\infty} + \left\|A RR^\top (\lbd_*-\lt)  \right\|_{\infty} + \varsigma \right)\\
\overset{\text{(\ref{eqn:rho:w:new:app})}}{=}  &-\rho_{w} \|\wh-\w_*\|_1 =-\rho_{w} \left(\|\wh_{\Obw}\|_1 + \|\wh_{\Ow} - \w_*\|_1\right).
\end{split}
\]

Substituting the above inequality into (\ref{eqn:3:new}), and the rest proof is identical to that of Lemma~\ref{lem:w}.

\section{Proof of Theorem~\ref{thm:main:2}}
The proof is similar to that of Theorem~\ref{thm:main}.  We introduce a different pseudo optimization problem
\[
\min_{\w \in \Omega}  \; g(\w) - \w^{\top}\Ah R^{\top}\lbd_*  + \gamma_w \|\w\|_1
\]
whose optimal solution is denoted by $\wt$. In the following, we will first bound the difference between $\wt$ and $\w_*$, and then $\lh$ and $\lbd_*$.

From the optimality of $\wt$ and $\w_*$, we derive the following lemma to bound their difference.
\begin{lem} \label{lem:ano:1}
Denote
\begin{equation} \label{eqn:ano:1}
 \rho_{w}= \left\| A ( I-RR^{\top})\lbd_* \right\|_\infty.
\end{equation}
By choosing $\gamma_{w} \geq 2 \rho_{w}$, we have
\[
\|\wt - \w_*\|_2 \leq \frac{3\gamma_w\sqrt{s_w}}{\alpha}, \ \|\wt - \w_*\|_1 \leq \frac{12\gamma_w s_w}{\alpha}, \textrm{ and } \frac{\|\wt - \w_*\|_1}{\|\wt - \w_*\|_2} \leq 4\sqrt{s_w}.
\]
\end{lem}

Based on the property of the random matrix $R$ described in Property~\ref{thm:jl}, we have the following lemma to bound $\rho_{w}$ in (\ref{eqn:ano:1}).
\begin{lem} \label{lem:ano:2}  With a probability at least $1 - \delta$, we have
\[
 \rho_{w}= \left\| A (I-RR^{\top})\lbd_* \right\|_\infty \leq   \|\lbd_*\|_2 \sqrt{\frac{c}{m}\log \frac{4d}{\delta}}
\]
provided (\ref{eqn:delta}) holds.
\end{lem}

Combining Lemma~\ref{lem:ano:1} with Lemma~\ref{lem:ano:2}, we immediately obtain the following lemma.
\begin{lem} \label{thm:ano:1}
Set
\[
\gamma_{w} \geq 2  \|\lbd_*\|_2 \sqrt{\frac{c}{m}\log \frac{4d}{\delta}}.
\]
With a probability at least $1-\delta$, we have
\[
\|\wt - \w_*\|_2 \leq \frac{3\gamma_w\sqrt{s_w}}{\alpha}, \ \|\wt - \w_*\|_1 \leq \frac{12\gamma_w s_w}{\alpha}, \textrm{ and } \frac{\|\wt - \w_*\|_1}{\|\wt - \w_*\|_2} \leq 4\sqrt{s_w}
\]
provided (\ref{eqn:delta}) holds.
\end{lem}

We then provide two lemmas that lead to Theorem~\ref{thm:main:2}.
\begin{lem} \label{lem:ano:3}
Denote
\begin{equation} \label{eqn:ano:2}
 \rho_{\lambda}= \left\|(RR^{\top} -I)A^{\top}\w_* \right\|_{\infty} + \left\|RR^{\top}A^{\top}(\wt - \w_*) \right\|_{\infty}.
\end{equation}
By choosing $\gamma_{\lambda} \geq 2 \rho_{\lambda}$, we have
\[
\|\lh - \lbd_*\|_2 \leq \frac{3\gamma_{\lambda} \sqrt{s_\lambda}}{\beta}, \ \|\lh - \lbd_*\|_1 \leq \frac{12\gamma_{\lambda} s_\lambda}{\beta}, \textrm{ and }  \frac{\|\lh - \lbd_*\|_1}{\|\lh - \lbd_*\|_2} \leq  4 \sqrt{ s_\lambda}.
\]
\end{lem}
\begin{lem} \label{lem:ano:4} Assume the conclusion in Lemma~\ref{thm:ano:1} happens.  With a probability at least $1 - 2 \delta$, we have
\[
 \rho_{\lambda}\leq\|A^{\top}\w_*\|_2 \sqrt{\frac{c}{m}\log \frac{4 n}{\delta}}+\frac{3\gamma_w\sqrt{s_w}}{\alpha} \left(1 + 7 \zeta_{A^\top}(16 s_w) \sqrt{\frac{c}{m} \left(\log \frac{4n}{\delta}+ 16s_w \log \frac{9 d}{8 s_w}\right)}\right)
\]
provided (\ref{eqn:delta}) holds.
\end{lem}

\subsection{Proof of Lemma~\ref{lem:ano:1}}
Recall the definitions of $\Ow$, $\Obw$ and $\G(\cdot)$ in Appendix~\ref{sec:lem:w}. We further define
\[
\Gt(\w) =g(\w) - \w^{\top}\Ah R^{\top}\lbd_*  + \gamma_w \|\w\|_1
\]

Let $\v \in \partial \|\w_*\|_1$ be any subgradient of  $ \|\cdot\|_1$ at $\w_*$. Then, we have
\[
\u = \nabla g(\w_*)- AR R^{\top}\lbd_* + \gamma_w  \v \in  \partial \Gt(\w_*).~\footnote{In the case that $g(\cdot)$ is non-smooth, $\nabla g(\w_*)$ refers to a subgradient of $g(\cdot)$ at $\w_*$. In particular, we choose the subgradient that satisfies (\ref{eqn:opt:new}). }
\]
Using the fact that $\wt$ minimizes $\Gt(\cdot)$ over the domain $\Omega$ and $g(\cdot)$ is $\alpha$-strongly convex, we have
\begin{equation} \label{eqn:ano:1:new}
\begin{split}
0& \geq \Gt(\wt)- \Gt(\w_*)  \geq  \langle \wt-\w_*,  \u \rangle + \frac{\alpha}{2}\|\wt-\w_*\|_2^2 \\
=&\left \langle \wt-\w_*, \nabla g(\w_*)- AR R^{\top}\lbd_* + \gamma_w  \v  \right\rangle + \frac{\alpha}{2}\|\wt-\w_*\|_2^2.\\
\end{split}
\end{equation}
By setting $v_i= \sgn(\widetilde{w}_i)$, $\forall i \in \Obw$, we have $\langle \wt_{\Obw} , \v_{\Obw} \rangle  = \|\wt_{\Obw}\|_1$.  As a result,
\begin{equation} \label{eqn:ano:2:new}
\langle \wt-\w_* , \v \rangle = \langle \wt_{\Obw}, \v_{\Obw}\rangle+ \langle \wt_{\Ow} - \w_* , \v_{\Ow} \rangle \geq \|\wt_{\Obw}\|_1 - \|\wt_{\Ow} - \w_*\|_1.
\end{equation}
Combining (\ref{eqn:ano:1:new}) with (\ref{eqn:ano:2:new}), we have
\begin{equation} \label{eqn:ano:3:new}
\left \langle \wt-\w_*, \nabla g(\w_*)- AR  R^{\top}\lbd_* \right\rangle+\frac{\alpha}{2}\|\wt-\w_*\|_2^2 + \gamma_w  \|\wt_{\Obw}\|_1 \leq  \gamma_w  \|\wt_{\Ow} - \w_*\|_1 .
\end{equation}
Furthermore, we have
\begin{equation} \label{eqn:ano:4:new}
\begin{split}
& \left \langle \wt-\w_*, \nabla g(\w_*)- AR  R^{\top}\lbd_* \right\rangle \\
= &  \left\langle \wt-\w_*, \nabla g(\w_*)-  A \lbd_* \right\rangle + \left \langle \wt-\w_*, A (I- RR^{\top})\lbd_* \right\rangle\\
\overset{\text{(\ref{eqn:opt:new})}}{\geq}  & -\|\wt-\w_*\|_1  \left\|A (I- RR^{\top})\lbd_* \right\|_{\infty} \\
\overset{\text{(\ref{eqn:ano:1})}}{=}  &-\rho_{w} \|\wt-\w_*\|_1 =-\rho_{w} \left(\|\wt_{\Obw}\|_1 + \|\wt_{\Ow} - \w_*\|_1\right).
\end{split}
\end{equation}

From  (\ref{eqn:ano:3:new}) and  (\ref{eqn:ano:4:new}), we have
\[
\frac{\alpha}{2}\|\wt-\w_*\|_2^2  + (\gamma_{w} - \rho_{w}) \|\wt_{\Obw}\|_1 \leq (\gamma_{w} + \rho_{w})\|\wt_{\Ow} - \w_*\|_1.
\]
The rest proof is identical to that of~Lemma~\ref{lem:w}.

\subsection{Proof of Lemma~\ref{lem:ano:2}}
Lemma~\ref{lem:ano:2} is a byproduct of the proof of Lemma~\ref{lem:rho:w} in Section~\ref{sec:lem:rho:w}. Specifically, the upper bound for $U_1$ leads to this lemma.

\subsection{Proof of Lemma~\ref{lem:ano:3}}
Recall the definitions of $\Ol$, $\Obl$ and $\L(\cdot)$ in Section~\ref{sec:lem:lambda}. We further define
\[
\Lh(\lbd)  =  -h(\lbd) - \gamma_{\lambda} \|\lbd\|_1 + \min\limits_{\w \in \Omega}   g(\w)  - \w^{\top} \Ah R^{\top}\lbd + \gamma_w \|\w\|_1.\\
\]

Let $\v \in \partial \|\lbd_*\|_1$ be any subgradient of  $\|\cdot\|_1$ at $\lbd_*$. Then, we have
\[
\u=-\nabla h(\lbd_*)-  RR^{\top}A^{\top}\wt- \gamma_{\lambda} \v  \in   \partial \Lh(\lbd_*).~\footnote{In the case that $h(\cdot)$ is non-smooth, $\nabla h(\lbd_*)$ refers to a subgradient of $h(\cdot)$ at $\lbd_*$. In particular, we choose the subgradient that satisfies (\ref{eqn:lem1:4}).}
\]
Using the fact that $\lh$ maximizes $\Lh(\cdot)$ over the domain $\Delta$ and $h(\cdot)$ is $\beta$-strongly convex, we have
\begin{equation} \label{eqn:ano:lem1:1}
\begin{split}
0& \geq \Lh(\lbd_*)- \Lh(\lh)  \geq  \langle-(\lh - \lbd_*), \u \rangle + \frac{\beta}{2}\|\lbd_* - \lh\|_2^2 \\
=&\left\langle\lh - \lbd_*, \nabla h(\lbd_*)+  RR^{\top}A^{\top}\wt + \gamma_{\lambda} \v \right\rangle + \frac{\beta}{2}\|\lbd_* - \lh\|_2^2 .\\
\end{split}
\end{equation}
By setting $v_i=  \sgn(\widehat{\lambda}_i)$, $\forall  i \in \Obl$, we have $\langle \lh_{\Obl} , \v_{\Obl} \rangle  = \|\lh_{\Obl}\|_1$. As a result,
\begin{equation} \label{eqn:ano:lem1:2}
\langle \lh - \lbd_*, \v \rangle = \langle \lh_{\Obl} , \v_{\Obl} \rangle+ \langle \lh_{\Ol}- \lbd_* , \v_{\Ol} \rangle \geq \|\lh_{\Obl}\|_1 - \|\lh_{\Ol} - \lbd_*\|_1.
\end{equation}
Combining (\ref{eqn:ano:lem1:1}) with (\ref{eqn:ano:lem1:2}), we have
\begin{equation} \label{eqn:ano:lem1:3}
\left \langle \lh - \lbd_*, \nabla h(\lbd_*) + RR^{\top}A^{\top}\wt \right\rangle+\frac{\beta}{2}\|\lbd_* - \lh\|_2^2 + \gamma_{\lambda}\|\lh_{\Obl}\|_1 \leq  \gamma_{\lambda}\|\lh_{\Ol} - \lbd_*\|_1 .
\end{equation}

Furthermore, we have
\begin{equation} \label{eqn:ano:lem1:5}
\begin{split}
& \left \langle \lh - \lbd_*, \nabla h(\lbd_*) + RR^\top A^{\top}\wt\right\rangle \\
= & \left\langle \lh - \lbd_*, \nabla h(\lbd_*) + A^{\top}\w_* \right\rangle +  \left \langle \lh - \lbd_*, (RR^{\top} -I)A^{\top}\w_* \right\rangle +  \left \langle \lh - \lbd_*, R R^\top A^{\top}(\wt-\w_*) \right\rangle \\
\overset{\text{(\ref{eqn:lem1:4})}}{\geq}  & -\|\lh - \lbd_*\|_1 \left(\left\|(RR^{\top} -I)A^{\top}\w_* \right\|_{\infty} + \left\|RR^{\top}A^{\top}(\wt - \w_*) \right\|_{\infty}\right)\\
\overset{\text{(\ref{eqn:ano:2})}}{=}  &  -\rho_{\lambda} \|\lh - \lbd_*\|_1=-\rho_{\lambda} \left( \|\lh_{\Obl}\|_1  + \|\lh_{\Ol} - \lbd_*\|_1\right).
\end{split}
\end{equation}

From  (\ref{eqn:ano:lem1:3}) and  (\ref{eqn:ano:lem1:5}), we have
\[
\frac{\beta}{2}\|\lh - \lbd_*\|_2^2 + (\gamma_{\lambda} - \rho_{\lambda})\|\lh_{\Obl}\|_1 \leq (\gamma_{\lambda} + \rho_{\lambda})\|\lh_{\Ol} - \lbd_*\|_1.
\]
The rest proof is identical to that of~Lemma~\ref{lem:lambda}.
\subsection{Proof of Lemma~\ref{lem:ano:4}}
The proof is similar to that of Lemma~\ref{lem:rho:w}. The first term $\|(RR^{\top} -I)A^{\top}\w_* \|_{\infty}$ can be upper bounded by Lemma~\ref{lem:rho:lambda}. Thus, we move to the second term and have
\[
\left\|RR^{\top}A^{\top}(\wt - \w_*) \right\|_{\infty} \leq \underbrace{\left\|A^{\top}(\wt - \w_*) \right\|_{\infty}}_{:=V_1} + \underbrace{\left\|( RR^\top - I)A^{\top}(\wt - \w_*) \right\|_{\infty}}_{:=V_2}
\]

\paragraph{Bounding $V_1$} From our assumption, we have
\[
\left\|A^{\top}(\wt - \w_*) \right\|_{\infty} \leq \max_{j\in[n]} \|A_{*j}\|_2 \|\wt - \w_*\|_2 \overset{\text{(\ref{eqn:upper:A2})}}{\leq} \|\wt - \w_*\|_2.
\]

\paragraph{Bounding $V_2$} Similar to the bounding of $U_3$ in Section~\ref{sec:lem:rho:w}, we will utilize the fact that $\wt - \w_*$ is approximately sparse when the conclusion in Lemma~\ref{thm:ano:1} happens.  Define
\[
\begin{split}
\K_{d,16 s_w} =  \left\{\x \in \R^d: \|\x\|_2 \leq 1, \|\x\|_1 \leq 4\sqrt{s_w} \right\}.
\end{split}
\]
When the conclusion in Lemma~\ref{thm:ano:1} holds, we have
\begin{equation} \label{eqn:sparse:belong:w}
\frac{\wt - \w_*}{\|\wt - \w_*\|_2} \in \K_{d,16 s_w}
\end{equation}
and thus
\[
V_2 = \|\wt - \w_*\|_2\left\|( RR^\top - I)A^{\top}  \frac{\wt - \w_*}{\|\wt - \w_*\|_2} \right\|_{\infty} \overset{\text{(\ref{eqn:sparse:belong:w})}}{\leq} \|\wt - \w_*\|_2  \underbrace{\sup_{\z \in \K_{d,16 s_w}} \left\|( RR^\top - I)A^{\top} \z \right\|_{\infty}}_{:=V_3}.
\]

Similar to Lemma~\ref{lem:conver}, we can utilize techniques of covering number to bound $V_3$.
\begin{lem} \label{lem:conver:2} With a probability at least $1-\delta$, we have
\[
\sup_{\z \in \K_{d,16 s_w}} \left\|( RR^\top - I)A^{\top} \z \right\|_{\infty}  \leq 2(2+\sqrt{2})  \zeta_{A^\top}(16 s_w) \sqrt{\frac{c}{m} \left(\log \frac{4n}{\delta}+ 16s_w \log \frac{9 d}{8 s_w}\right)}.
\]
\end{lem}
Putting everything together, we have
\[
\begin{split}
 \rho_{\lambda} \leq &  \|A^\top \w_*\|_2 \sqrt{\frac{c}{m}\log \frac{4 n}{\delta}} +  \|\wt - \w_*\|_2 \left(1+2(2+\sqrt{2})  \zeta_{A^\top}(16 s_w) \sqrt{\frac{c}{m} \left(\log \frac{4n}{\delta}+ 16s_w \log \frac{9 d}{8 s_w}\right)} \right) \\
\leq & \|A^{\top}\w_*\|_2 \sqrt{\frac{c}{m}\log \frac{4 n}{\delta}}+\frac{3\gamma_w\sqrt{s_w}}{\alpha} \left(1 + 7 \zeta_{A^\top}(16 s_w) \sqrt{\frac{c}{m} \left(\log \frac{4n}{\delta}+ 16s_w \log \frac{9 d}{8 s_w}\right)}\right).
\end{split}
\]
\subsection{Proof of Lemma~\ref{lem:conver:2}}
Define
\[
\S_{d,16s_w}=  \left\{\x \in \R^d: \|\x\|_2 \leq 1, \|\x\|_0 \leq 16 s_w \right\} .
\]
Let $\S_{d,16s_\lambda}(\epsilon)$ be a $\epsilon$-net of $\S_{d,16s_\lambda}$ with smallest cardinality.
Following the proof of Lemma~\ref{lem:conver}, we have
\[
\begin{split}
 \sup_{\z \in \K_{d,16 s_w}} \left\|( RR^\top - I)A^{\top} \z \right\|_{\infty} \leq 2(2+\sqrt{2}) \underbrace{\sup\left\{\left\|( RR^\top - I)A^{\top} \z \right\|_{\infty}: \z \in \S_{d,16s_w}(1/2)\right\}}_{\kappa(1/2)}
 \end{split}
\]
where
\[
    \log |\S_{d,16s_w}(1/2)| \leq 16s_w \log\left( \frac{9 d}{8s_w}\right).
\]

We proceed by providing an upper bound for $\kappa(1/2)$. Following the proof of Lemma~\ref{lem:rho:lambda}, for each $\z \in \S_{d,16s_w}(1/2)$, we have with a probability at least $1-\delta$,
\[
\begin{split}
\left\| ( RR^\top - I)A^{\top} \z \right\|_\infty  \leq & \|A^{\top}\z\|_2 \sqrt{\frac{c}{m}\log \frac{4 n}{\delta}} \\
\overset{\text{(\ref{eqn:cs:upper})}}{\leq} & \zeta_{A^\top}(16 s_w) \|\z\|_2 \sqrt{\frac{c}{m}\log \frac{4 n}{\delta}} \leq \zeta_{A^\top}(16 s_w)  \sqrt{\frac{c}{m}\log \frac{4 n}{\delta}}.
\end{split}
\]
We complete the proof by taking the union bound over all $\z \in \S_{d,16s_w}(1/2)$.

\section{Proof of Theorem~\ref{thm:approximate}}
The proof is almost identical to that of Theorem~\ref{thm:main}. We just need to replace Lemmas~\ref{lem:lambda} and \ref{lem:w} with the following ones.

\begin{lem} \label{lem:lambda:app}
Denote
\begin{equation}  \label{eqn:rho:lambda:app}
 \rho_{\lambda}= \left\|(RR^\top -I)A^{\top}\w_* \right\|_{\infty}+(1+\mu)\tau.
\end{equation}
By choosing $\gamma_{\lambda} \geq 2 \rho_{\lambda}$, we have
\[
\|\lt - \lbd_*\|_2 \leq \frac{3\gamma_\lambda\sqrt{s_\lambda}}{\beta}, \ \|\lt - \lbd_*\|_1 \leq \frac{12\gamma_\lambda s_\lambda}{\beta}, \textrm{ and } \frac{\|\lt - \lbd_*\|_1}{\|\lt - \lbd_*\|_2} \leq 4\sqrt{s_\lambda}.
\]
\end{lem}
\begin{lem} \label{lem:w:app}
Denote
\begin{equation} \label{eqn:rho:w:app}
 \rho_{w}= \left\| A \left( I-RR^{\top}\right)\lbd_* \right\|_\infty + \left\|A RR^\top (\lbd_*-\lt)  \right\|_{\infty}+(1+\mu)\tau.
\end{equation}
By choosing $\gamma_{w} \geq 2 \rho_{w}$, we have
\[
\|\wh - \w_*\|_2 \leq \frac{3\gamma_w\sqrt{s_w}}{\alpha}, \ \|\wh - \w_*\|_1 \leq \frac{12\gamma_w s_w}{\alpha}, \textrm{ and } \frac{\|\wh - \w_*\|_1}{\|\wh - \w_*\|_2} \leq 4\sqrt{s_w}.
\]
\end{lem}

\subsection{Proof of Lemma~\ref{lem:lambda:app}}
Recall the definition of $\L(\cdot)$ in Section~\ref{sec:lem:lambda}. From the fact that the optimal solution $\lbd_*'$ lies in the interior of $\Delta$, we have
\begin{equation} \label{eqn:lem1:4:app}
 \nabla \L(\lbd_*')= -\nabla h(\lbd_*') - A^\top \w_*' =  0.
\end{equation}
Then,
\begin{equation} \label{eqn:lem1:5:app}
\begin{split}
& \left \langle \lt - \lbd_*, \nabla h(\lbd_*) + RR^\top A^{\top}\w_*\right\rangle \\
= & \left\langle \lt - \lbd_*, \nabla h(\lbd_*') + A^{\top}\w_*' \right\rangle + \left \langle \lt - \lbd_*, \nabla h(\lbd_*) -\nabla h(\lbd_*') \right \rangle \\
&+\left \langle \lt - \lbd_*,  A^{\top}\w_*-A^{\top}\w_*'\right\rangle+\left \langle \lt - \lbd_*, (RR^{\top} -I)A^{\top}\w_* \right\rangle\\
\overset{\text{(\ref{eqn:lem1:4:app})}}{\geq}  & -\|\lt - \lbd_*\|_1 \left(\|\nabla h(\lbd_*) -\nabla h(\lbd_*')\|_{\infty}+\left\|A^{\top}\w_*-A^{\top}\w_*'\right\|_{\infty}+\left\|(RR^{\top} -I)A^{\top}\w_* \right\|_{\infty} \right).
\end{split}
\end{equation}

From our assumptions, we have
\begin{align}
 \|\nabla h(\lbd_*) -\nabla h(\lbd_*')\|_\infty &\leq \|\nabla h(\lbd_*) -\nabla h(\lbd_*')\|_2  \overset{\text{(\ref{eqn:smooth:h})}}{\leq}  \mu \|\lbd_*-\lbd_*'\|_2 \overset{\text{(\ref{eqn:tau})}}{\leq}  \mu \tau, \label{eqn:appr:1}\\
 \left\| A^{\top}(\w_*-\w_*') \right\|_\infty &\leq \max_{j\in[n]} \|A_{*j}\|_2 \|\w_*-\w_*'\|_2 \overset{\text{(\ref{eqn:upper:A2})}}{\leq} \|\w_*-\w_*'\|_2   \overset{\text{(\ref{eqn:tau})}}{\leq}   \tau.\label{eqn:appr:2}
\end{align}
From (\ref{eqn:lem1:5:app}), (\ref{eqn:appr:1}) and (\ref{eqn:appr:2}), we have
\[
\begin{split}
& \left \langle \lt - \lbd_*, \nabla h(\lbd_*) + RR^\top A^{\top}\w_*\right\rangle \\
\geq  &  -\|\lt - \lbd_*\|_1\left(\left\|(RR^{\top} -I)A^{\top}\w_* \right\|_{\infty}+(1+\mu)\tau \right)\\
\overset{\text{(\ref{eqn:rho:lambda:app})}}{=}  &  -\rho_{\lambda} \|\lt - \lbd_*\|_1=-\rho_{\lambda} \left( \|\lt_{\Obl}\|_1  + \|\lt_{\Ol} - \lbd_*\|_1\right).
\end{split}
\]
Substituting the above inequality into (\ref{eqn:lem1:3}), and the rest proof is identical to that of Lemma~\ref{lem:lambda}.
\subsection{Proof of Lemma~\ref{lem:w:app}}
Recall the definition of $\G(\cdot)$ in Appendix~\ref{sec:lem:w}. From the fact that the optimal solution $\w_*'$ lies in the interior of $\Omega$, we have
\begin{equation} \label{eqn:opt:new:appr}
 \nabla \G(\w_*')=  \nabla g(\w_*') - A \lbd_*' = 0.
\end{equation}
Then,
\begin{equation} \label{eqn:4:new:app}
\begin{split}
& \left \langle \wh-\w_*, \nabla g(\w_*)- ARR^{\top} \lt \right\rangle \\
= & \left \langle \wh-\w_*, \nabla g(\w_*') - A \lbd_*' \right \rangle + \langle \wh-\w_*, \nabla g(\w_*) - \nabla g(\w_*') \rangle + \left \langle \wh-\w_*,  A \lbd_*' - A \lbd_* \right \rangle \\
&+\left \langle \wh-\w_*, A (I- RR^{\top})\lbd_* \right\rangle +  \left \langle \wh-\w_*, A RR^\top (\lbd_*-\lt) \right\rangle\\\
\overset{\text{(\ref{eqn:opt:new:appr})}}{\geq}  & -\|\wh-\w_*\|_1  \left( \left\|\nabla g(\w_*) - \nabla g(\w_*')  \right\|_{\infty} + \left\|A \lbd_*' - A \lbd_* \right\|_{\infty} \right)\\
&-\|\wh-\w_*\|_1  \left(\left\|A (I- RR^{\top})\lbd_* \right\|_{\infty} + \left\|A RR^\top (\lbd_*-\lt)  \right\|_{\infty}\right).
\end{split}
\end{equation}

From our assumptions, we have
\begin{align}
 \|\nabla g(\w_*)-  \nabla g(\w_*')\|_\infty &\leq \|\nabla g(\w_*)-  \nabla g(\w_*')\|_2  \overset{\text{(\ref{eqn:smooth:g})}}{\leq}  \mu \|\w_*-\w_*'\|_2 \overset{\text{(\ref{eqn:tau})}}{\leq}  \mu \tau, \label{eqn:appr:1:new}\\
\|A (\lbd_*'- \lbd_*)\|_\infty &\leq \max_{i\in[d]} \|A_{i*}\|_2 \|\lbd_*'- \lbd_*\|_2 \overset{\text{(\ref{eqn:upper:A1})}}{\leq} \|\lbd_*'- \lbd_*\|_2 \overset{\text{(\ref{eqn:tau})}}{\leq} \tau.\label{eqn:appr:2:new}
\end{align}
From (\ref{eqn:4:new:app}), (\ref{eqn:appr:1:new}) and (\ref{eqn:appr:2:new}), we have
\[
\begin{split}
& \left \langle \wh-\w_*, \nabla g(\w_*)- ARR^{\top} \lt \right\rangle \\
\geq & -\|\wh-\w_*\|_1 \left(\left\|A (I- RR^{\top})\lbd_* \right\|_{\infty} + \left\|A RR^\top (\lbd_*-\lt)  \right\|_{\infty}  + (1  + \mu)  \tau \right)\\
\overset{\text{(\ref{eqn:rho:w:app})}}{=}  &-\rho_{w} \|\wh-\w_*\|_1 =-\rho_{w} \left(\|\wh_{\Obw}\|_1 + \|\wh_{\Ow} - \w_*\|_1\right).
\end{split}
\]
Substituting the above inequality into (\ref{eqn:3:new}), and the rest proof is identical to that of Lemma~\ref{lem:w}.
\end{document}